%% file: acl_latex.tex
\documentclass[11pt]{article}

\usepackage[final]{acl}

\usepackage{times}
\usepackage{latexsym}
\usepackage{booktabs}
\usepackage{multirow}
\usepackage{subcaption}
\usepackage{amsmath}
\usepackage{amsfonts}
\usepackage{tcolorbox}
\usepackage{fancyvrb}
\usepackage{fvextra}
\usepackage[table]{xcolor}
\definecolor{hl}{RGB}{222,235,250}
\usepackage{listings}
\usepackage{listings}
\usepackage{multirow}
\lstdefinestyle{prompt}{basicstyle=\ttfamily\scriptsize, breaklines=true,
  breakatwhitespace=true, breakindent=0pt, columns=fullflexible, keepspaces=true,
  frame=single, framesep=3pt, xleftmargin=3pt, xrightmargin=3pt,
  aboveskip=6pt, belowskip=8pt, captionpos=t}
\DeclareCaptionFormat{promptttl}{\textbf{#3}}   
\lstnewenvironment{prompt}[1]
  {\lstset{style=prompt, title={\small\normalfont\bfseries #1}}}
  {}
\usepackage[T1]{fontenc}

\usepackage[utf8]{inputenc}

\usepackage{microtype}

\usepackage{inconsolata}

\usepackage{graphicx}

\usepackage{enumitem}
\setlist[description]{style=unboxed, leftmargin=0pt}

\usepackage{xcolor}         

\title{Reasoning Consensus: Auditable Chain-of-thought Ensembling via Weighted DAG Aggregation}

\author{\textbf{Amruta Parulekar}, \textbf{Jinu Lee}, \textbf{Dilek Hakkani-T\"ur}, \textbf{Hari Sundaram} \\
  University of Illinois Urbana-Champaign \\
  \texttt{\{amp20, jinulee2, dilek, hs1\}@illinois.edu} \\}

\begin{document}
\maketitle

\begin{abstract}

Large Language Models (LLMs) reason in free-form chain-of-thought (CoT) without revealing which steps were well supported and which alternatives were rejected. In high-stakes domains, users then cannot judge if an answer rests on sound reasoning. Majority voting and multi-agent debate improve answers using model agreement, but do not expose agreement on individual steps.
\textsc{Reasoning Consensus} (RC) extracts a typed reasoning graph from each CoT, merges equivalent steps across graphs from diverse models, and weights every step by how many chains attest to it.
RC reveals this step-wise support and the varied justifications behind each answer.
On 7 benchmarks across law, science, medicine, narrative reasoning and logic, RC matches majority voting and debate in accuracy (69.4\% vs.\ 69.0--70.3\%).
An LLM judge prefers RC's best-supported graph to other graphs for the same answer. In a user study, participants read RC graphs faster than debate transcripts or vote tallies, and 14 of 18 preferred them for future use.\footnote{Our code is available at \href{https://anonymous.4open.science/r/RC-17C8/}{this link}}



\end{abstract}


\input{section/01_introduction.tex}

\input{section/02_related.tex}

\input{section/03_methodology.tex}
\input{section/04_setup.tex}

\input{section/05_results.tex}

\input{section/08_limitations.tex}

\input{section/A1_prompts.tex}

\input{section/A3_graphaccuracies.tex}
\input{section/A4_nodetypes.tex}

\end{document}

%% file: section/01_introduction.tex
\vspace{-9pt}

\section{Introduction}
\label{sec:intro}
\vspace{-3pt}

\begin{figure*}[t]
  \centering
  \includegraphics[width=\textwidth]{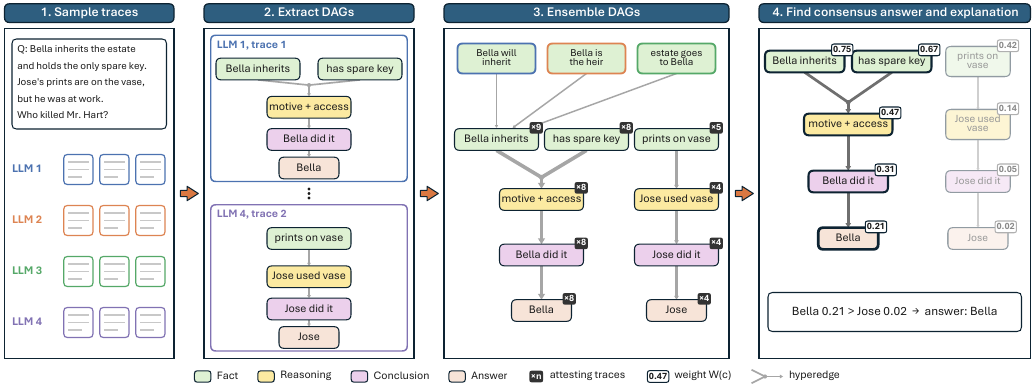}
   \vspace{-20pt}
  \caption{\textbf{Overview of \textsc{Reasoning Consensus}}. (1) Sample traces from different models. (2) Extract a DAG of typed steps from each trace. (3) Merge equivalent steps across traces, counting the traces that \textit{attest} to each ($\times n$). (4) Weight each step by its attestation and its premises' support; return the consensus and competing subgraphs.}
  \vspace{-12pt}
  \label{fig:pipeline}
\end{figure*}


Large language models (LLMs) are increasingly used to take decisions, including in high-stakes domains such as law, medicine, and science. LLMs usually reason through a single free-form chain-of-thought (CoT) \citep{wei2023chainofthoughtpromptingelicitsreasoning}, though complex problems often have several possible solutions \citep{yang2026accuracyevaluatingstrategydiversity}. Linear CoTs do not show which reasoning steps were well supported, and which alternatives were rejected or weakly supported. A single linear CoT can have incorrect steps leading to brittle reasoning \cite{zeng2026halluguarddemystifyingdatadrivenreasoningdriven} or the right answer for the wrong reasons \citep{Dahl_2024, lee_evaluating_2026}. Thus, \textit{a framework for exposing the LLM reasoning structure is crucial for improved transparency and safe use of LLMs.}


Agreement between independent reasoners can expose brittle reasoning. An answer reached by several independent CoTs is more likely to be correct \citep{wang2023selfconsistencyimproveschainthought}, and content that samples disagree on is more likely to be hallucinated  \citep{manakul2023selfcheckgptzeroresourceblackboxhallucination}. Existing methods to combine diverse reasoning either vote over final answers \citep{naik2024diversitythoughtimprovesreasoning}, select a single chain \citep{xu-etal-2025-collaborative, park2024ensemblinglargelanguagemodels}, or make models discuss their solutions \citep{wang2024mixtureofagentsenhanceslargelanguage,
chen2024reconcileroundtableconferenceimproves}. However, these methods do not reveal the degree of agreement among models on the reasoning steps, or the lines of reasoning that led to different answers.  ARGORA
\citep{jin2026argoraorchestratedargumentationcausally} does this using interactive debate graphs, but models tend to favor agreement in debate, switching answers instead of challenging flawed reasoning \cite{wynn2025talkisntcheapunderstanding}.

Our framework, \textsc{Reasoning Consensus} (\textsc{RC}), is designed to be auditable and transparent~\cite{Felzmann2020TowardsTB}, and to reveal the reasoning structure behind each answer \citep{article}. RC gives higher weight to steps that most traces state, and errors in a few traces (a hallucinated fact, a miscalculation) receive lower weight. The \textit{key insight} is that independent CoT traces that agree on the steps behind an answer give stronger evidence than traces that agree on the answer alone. 



We evaluated RC on 7 benchmarks in law, science, medicine, narrative reasoning and logic, with traces from 4 open model families. Our experiments show that RC matches majority voting (MV) and multi-agent debate in accuracy (69.4\% against 69.0--70.3\%), while using 4.1 times the tokens of MV. On tied votes, where MV picks an answer at random, RC is right 8.6 points more often than MV. On expert-knowledge questions, RC alone answers 58 questions correctly and MV alone 33 ($p{=}0.011$). Across all benchmarks, with traces from a single model, RC alone answers 101 correctly and self-consistency \citep{wang2023selfconsistencyimproveschainthought} alone 72 ($p{=}0.03$). At equal length, a blinded LLM judge prefers the consensus subgraph (the best-supported reasoning for the chosen answer) to another subgraph for the same answer, chosen at random, with probability 0.61 (95\% credible interval 0.58--0.64). In a study with 18 participants, people found the central step behind the majority answer, and the step where two answers diverge, as accurately with RC's subgraphs for the two answers as with vote tallies or debate transcripts (82\%, 82\% and 79\%). With RC's subgraphs they needed a median of 3.4 minutes per question, against 5.1 with vote tallies and 4.4 with debate transcripts, and 14 of 18 preferred RC's subgraphs for future use. We make two contributions:

\begin{description}
  \item[A framework for consensus LLM reasoning:] We develop RC, which merges independent reasoning traces from several LLMs into one graph. In contrast, MV keeps only the final answers \citep{wang2023selfconsistencyimproveschainthought}, and in debate, ARGORA included, models defer to their peers to reach agreement. RC merges equivalent steps by embedding clustering with LLM confirmation, and weights each step by its attestation count (the number of traces stating it) and the weights of its premises. RC thus measures every step's support without labels or training.
  \item[Auditable and transparent by design:] We design RC to reveal the reasoning behind each answer: the consensus subgraph, the best-supported subgraph for each competing answer, and the support for every step. In contrast, MV reports only a tally of answers \citep{wang2023selfconsistencyimproveschainthought} and debate a long transcript \citep{chen2024reconcileroundtableconferenceimproves}; in both, the reader has to find where two answers' reasoning diverges. RC builds each subgraph by following the best-supported premises of every step back from the answer. Comparing two answers' subgraphs then shows where their reasoning parts. A reader can thus audit an answer by checking the graph.
\end{description}


%% file: section/02_related.tex



\section{Related Work}
\label{sec:related}
\vspace{-4pt}



\paragraph{Structured reasoning with LLMs.} Structured reasoning organizes the reasoning trace into explicit structures, such as trees \citep{yao2023treethoughtsdeliberateproblem} and graphs \citep{Besta_2024}. These structures enable principled exploration of the solution space by tracking previous reasoning paths and failures (backtracking), and also increase reasoning monitorability by providing structural cues. Recent works extend structured reasoning by combining multiple LLMs \citep{wang2024mixtureofagentsenhanceslargelanguage, park2024ensemblinglargelanguagemodels, xu-etal-2025-collaborative, yang2025multillmcollaborativesearchcomplex}, further promoting reasoning path diversity. However, these approaches intentionally collapse these paths into a single solution, discarding alternatives that can measure certainty or serve as alternative explanations when the user disagrees with the LLM.

\vspace{-2pt}

\paragraph{Multi-agent debate.} Multi-agent debate methods like Mixture-of-Agents \citep{wang2024mixtureofagentsenhanceslargelanguage}
and ReConcile \citep{chen2024reconcileroundtableconferenceimproves} use conversation between different LLMs to increase diversity. Multi-agent debate produces a large amount of unstructured text, reducing the monitorability of the reasoning process.  To preserve initial argument diversity in a multi-agent setup, ARGORA \citep{jin2026argoraorchestratedargumentationcausally} generates argumentation graphs from multi-expert discussions. It tests the weakness of competing claims through causal intervention, but relies on constrained interactive debate. However, multi-agent debate introduces peer pressure to agree between agents, causing errors to spread easily \citep{wynn2025talkisntcheapunderstanding, kasprova2026politedisagreeunderstandingsycophancy}. 

 In contrast with these works, RC extracts graphs from independent and unconstrained
traces, weighting each step by how many traces attest to it. On disagreement, both answers are retained with their supporting steps, making every step auditable. \textsc{RC}
keeps the full weighted graph of attested steps and never forces models to agree, thus promoting both correctness and monitorability of reasoning.



%% file: section/03_methodology.tex
\section{Methodology}
\label{method}

Our pipeline (Figure~\ref{fig:pipeline}) has five stages: reasoning trace
generation (\S\ref{trace}), DAG extraction (\S\ref{dag}), hypergraph construction via clustering (\S\ref{node}), ensemble weighting (\S\ref{ens}) and finding
the consensus (\S\ref{consens}). Table~\ref{tab:notation} summarizes the
notation used throughout this section. Prompts, step-wise details and step-wise analyses in Appendices ~\ref{app:prompts}, \ref{app:method}, \ref{analyses}.

\begin{table}[htbp]
\centering
\small
\begin{tabular}{@{}l p{4.9cm}@{}}
\toprule
\textbf{Symbol} & \textbf{Meaning} \\
\midrule
$\mathcal{T}$ & set of reasoning traces for a query \\
$G_t = (V_t, E_t)$ & typed DAG of trace $t$ \\
$P(v)$ & premise set of step $v$ \\
$\pi: V \to C$ & clustering that maps steps into clusters \\
$HG_{ens} = (C, H)$ & merged ensemble DAHG \\
$\alpha(c)$ & cluster attestation $\in [0,1]$ \\
$\phi(h)$ & hyperedge strength $\in [0,1]$ \\
$W(c)$ & cluster strength $\in [0,1]$ \\
$a_y$; $\hat{y}$ & answer node (label $y$); consensus ans. \\
$S(c)$ & proof subgraph of $HG_{ens}$ rooted at $c$ \\
\bottomrule
\end{tabular}
\caption{Notation used in Section~\ref{method}.}
\vspace{-12pt}
\label{tab:notation}
\end{table}

\subsection{Reasoning Trace Generation}
\label{trace}

For every query, we sample $K$ reasoning traces from $N$ different models to obtain a set $\mathcal{T}$ of $NK$ traces. Each model is constrained to a fixed schema requiring step-by-step reasoning and a final verdict.
\subsection{DAG Extraction}
\label{dag}
 
An LLM extractor converts the reasoning trace $t$ into a DAG of steps $G_t$ connecting each step to its logical premises. Based on the taxonomy in
\citealp{lee2025reasoningflowsemanticstructurecomplex}, each step is labeled either
\emph{Planning} (what to do next), \emph{Fact} (facts that are commonsense or stated in the query), \emph{Reasoning} (intermediate
inference), or \emph{Conclusion} (model's last verdict, attached to the automatically parsed \emph{Answer} node).
The predecessors of a step $v$ form a \textit{premise set} $P(v)$, which denotes the premises that collectively support $v$.
 
\subsection{Ensemble Hypergraph Construction}
\label{node}
 
\paragraph{Clustering steps.} We cluster semantically equivalent steps that appear across different traces. Steps are first
embedded using a sentence embedding model. These steps are then clustered by average-linkage agglomerative clustering on cosine
distance, i.e., starting from individual clusters, two clusters are merged only when the mean cosine similarity between their
members is at least $\tau = 0.55$. Finally, we check whether the clusters are actually semantically equivalent using an LLM as a judge.

\paragraph{Clustering premise sets into hyperedges.} Let
clustering $\pi: V \to C$ be the map from every step to its cluster. Then, if steps and their premise sets from two different traces all map to the same cluster, i.e., $\pi(v_1)=\pi(v_2)$ and $\pi(P(v_1)) = \pi(P(v_2))$, we can merge these premise sets into a single \emph{premise hyperedge} $h: \pi(P(v_1)) \rightarrow \pi(v_1)$ that collectively connects a cluster to the supporting premise cluster set. If $\pi(v_1)=\pi(v_2)$ but $\pi(P(v_1)) \neq \pi(P(v_2))$, we keep two separate hyperedges $h_1: \pi(P(v_1)) \rightarrow \pi(v_1)$ and $h_2: \pi(P(v_2)) \rightarrow \pi(v_1)$. Finally, we break cycles into separate subgraphs. This defines the \emph{ensemble directed acyclic hypergraph (DAHG)} $HG_{ens}$ of clusters $C$ and premise hyperedges $H$. We also track traces that attest to clusters and hyperedges.
 
\subsection{Weight Assignment}
\label{ens}

We recursively compute \emph{weight} $W(c)$ of each cluster $c$ in the ensemble DAHG $HG_{ens}=(C, H)$.

For leaf step clusters (\textit{Fact}) without premise sets, the weight is simply the \emph{attestation} $\alpha(c)$, or the ratio of traces that attest to the cluster $c$. For instance, if the cluster contains steps from 8 out of 12 traces, the attestation becomes $\frac{8}{12} \approx 0.67$.

For non-leaf clusters, a step's correctness can be gauged from both the frequency of the step and the frequency of the premise set that leads to the step. Even if a step appears frequently, highly scattered premise hyperedges indicate spurious reasoning instead of following a standard path. Therefore, we define the hyperedge's \emph{strength} $\phi(h)$ as the average weight of the premise clusters, and define the weight as the product of the strongest hyperedge's strength and the cluster's attestation.
\begin{equation}
W(c) = \alpha(c)\,\max_{h_{: P\rightarrow c} \in H} \phi(h),
\label{eq:score}
\end{equation}where
\begin{equation}
\phi(h_{: P\rightarrow c}) = \frac{1}{|P|} \sum_{p \in P} W(p).
\label{eq:phi}
\end{equation}
\newpage
For example, a cluster $c$ attested by 8 out of 12 traces with hyperedges
$\{h_1, h_2\}$, with $\phi(h_1) = 0.5$ and
$\phi(h_2) = 0.3$, weighs
$W(c) = \tfrac{8}{12} \cdot 0.5 \approx 0.33$. Note that the computation involves alternative recursion: cluster weights require hyperedge strengths, which are derived from other clusters' weights. So we compute the step weights and hyperedge strength following the topological order.
 


 
\subsection{Finding the Consensus}
\label{consens}
 
The consensus answer $\hat{y}$ is the label whose answer node $a_y$ has highest weight,
\begin{equation}
\hat{y} = \arg\max_{y} W(a_y).
\end{equation}
Ties are broken by majority vote followed by random draw. Since every node's weight is scaled by $\alpha(c)$, an
answer's weight reflects not only how many traces reach it but also how many
share the supporting steps. 
This helps us favor answers supported by the same reasoning across traces over  coincidentally frequent answers. Attenuation could favor shallow traces, but depth is not a systematic bias as competing answers are reached at similar
depth on average (mean gap $0.2$ levels, s.d.\ $1.0$).
 
\paragraph{Consensus subgraph.} To give the best explanation for an answer $a_y$, we define a \emph{consensus subgraph} $S(a_{\hat{y}})$. At each cluster we take its strongest
hyperedge, $\arg\max_{h_{:P\rightarrow c} \in H} \phi(h)$, and expand its members until hitting the leaf clusters. Choosing other hyperedges than the strongest one
gives alternative justifications of the same answer. Similar greedy unfolding
from other answers gives most supported reasoning for each
competing answer. 
 
\subsection{Summary}
 
Reasoning Consensus (RC) turns unstructured reasoning traces from diverse models into one
ensemble DAHG. The merging process assigns weights to steps by how often that step or its explanation appears in individual traces, while preserving each trace's reasoning structure. This ensemble DAHG can be used to choose the best explanation for an answer, or compare the attestation strength of different reasoning paths. We next describe benchmarks, models, and baselines used to evaluate RC.
 

%% file: section/04_setup.tex
\section{Experimental Setup}
\label{sec:setup}

\paragraph{Datasets.} We chose seven benchmarks covering three different kinds of
reasoning to test RC.
\newpage
\emph{Expert knowledge application}: \textbf{SARA}
\citep{holzenberger2020datasetstatutoryreasoningtax} (272 questions) applies US
tax statutes to legal cases, \textbf{GPQA Diamond} \citep{rein2023gpqagraduatelevelgoogleproofqa}
(198) tests graduate-level science, and \textbf{MedBullets}
\citep{chen2026benchmarkinglargelanguagemodels} (190 USMLE-style cases\footnote{We use the 190
questions that do not refer to a figure.}) tests clinical medicine. Here
single traces can make reasoning errors, and rule-based problems
decompose into conjunctions and alternatives reflecting the structure
our merge preserves. \emph{Narrative evidence weighing}: the three
\textbf{MuSR} domains \citep{sprague2024musrtestinglimitschainofthought},
Murder Mysteries (MM, 250), Object Placements (OP, 256) and Team Allocation
(TA, 250), require long multi-hop chains over a story, where different reasoning supports different answers. \emph{Formal entailment}: \textbf{FOLIO}
\citep{han2024folionaturallanguagereasoning} (203) asks if a conclusion is
true, false or \emph{uncertain} given premises, testing if
weak support can be differentiated from uncertainty.

\paragraph{Models.} We use 4 open instruction-tuned models from different
families to generate traces. Gemma-4-E4B \citep{gemmateam2026gemma4technicalreport}, Qwen3-8B
\citep{yang2025qwen3technicalreport}, OLMo-3-7B \citep{olmo2026olmo3} and Phi-4
\citep{abdin2024phi4technicalreport}. Each
samples $K{=}3$ traces per question (temperature 0.6, top-$p$ 0.95), giving 12
 total traces. We use Qwen3.8-27B to extract DAGs (temperature 0.3) and
verify merges, and All-mpnet-base-v2 \citep{reimers-2019-sentence-bert} to embed steps for clustering. 
 All runs used vLLM \citep{kwon_efficient_2023} on NVIDIA A100 GPUs.

\begin{table*}[t]
\centering
\scriptsize
\renewcommand{\arraystretch}{1.12}
\setlength{\tabcolsep}{3pt}
\resizebox{\textwidth}{!}{%
\begin{tabular}{llccccccccc}
\toprule
 & & \multicolumn{3}{c}{\textit{Expert knowledge}} & \multicolumn{3}{c}{\textit{Narrative evidence}} & \textit{Formal} & & \\
\cmidrule(lr){3-5}\cmidrule(lr){6-8}\cmidrule(lr){9-9}
\multicolumn{2}{l}{\textbf{Method}} & \textbf{SARA} & \textbf{GPQA-D} & \textbf{MedB.} & \textbf{MuSR-MM} & \textbf{MuSR-OP} & \textbf{MuSR-TA} & \textbf{FOLIO} & \textbf{Wtd. Avg} & \textbf{Tok/q} \\
\midrule
\multicolumn{11}{l}{\textit{(a) Four-model ensemble, 12 generations per question; mean (SD) over 3 seeds}} \\
\midrule
\multicolumn{2}{l}{RC (ours)}
  & \textbf{86.8} (0.4) & 58.4 (2.6) & 69.6 (1.1)
  & 66.5 (1.6) & \textbf{52.3} (1.4) & 67.3 (0.6)
  & 84.4 (2.0) & 69.4 (0.7) & 65.1k \\
\multicolumn{2}{l}{Majority vote (Mixed SC)}
  & \underline{85.9} (0.2) & 57.2 (1.9) & 67.7 (1.2)
  & 66.7 (0.6) & \underline{52.0} (1.7)
  & \underline{67.6} (0.7)
  & 85.4 (0.3) & 69.0 (0.4) & 15.8k \\
\multicolumn{2}{l}{Mixture-of-Agents}
  & 84.3 (2.2) & \textbf{63.1} (3.5) & \textbf{73.3} (2.6)
  & \textbf{72.1} (3.6) & 47.1 (2.3)
  & \underline{67.6} (3.4)
  & \textbf{86.2} (0.9) & \textbf{70.3} (1.7) & 31.6k \\
\multicolumn{2}{l}{ReConcile}
  & 84.8 (2.2) & \underline{60.3} (1.1)
  & \underline{72.6} (0.0)
  & \underline{69.5} (1.9) & 46.5 (1.4)
  & \textbf{68.5} (1.6)
  & \underline{85.7} (1.3) & \underline{69.5} (0.2)
  & 33.6k \\
\midrule
\multicolumn{11}{l}{\textit{(b) Same-model pools: 12 traces of one model vs.\ the mixed pool (seed 1234)}} \\
\midrule
\multirow{2}{*}{Mixed (4 models)} & RC & \textbf{86.8} & \textbf{60.6} & \cellcolor{hl}\textbf{70.5} & \textbf{66.8} & 52.7 & \textbf{67.2} & 84.7 & \cellcolor{hl}\textbf{69.9} & \\
 & MV (SC) & 85.7 & 58.1 & 66.3 & 66.0 & \textbf{53.1} & 66.8 & \textbf{85.2} & 68.9 & \\
\midrule
\multirow{3}{*}{Gemma-4-E4B $\times$12} & RC & \cellcolor{hl}\textbf{87.9} & \cellcolor{hl}\textbf{63.6} & 63.2 & 66.0 & 43.0 & \cellcolor{hl}\textbf{71.6} & \cellcolor{hl}\textbf{87.2} & \textbf{68.9} & \\
 & MV (SC) & 87.5 & 60.1 & 62.6 & 66.0 & 43.0 & \textbf{71.6} & 86.7 & 68.3 & \\
\midrule
\multirow{3}{*}{Phi-4 $\times$12} & RC & \textbf{84.6} & \textbf{59.6} & \cellcolor{hl}\textbf{70.5} & \cellcolor{hl}\textbf{68.8} & 46.5 & \textbf{65.2} & \textbf{84.2} & \textbf{68.4} & \\
 & MV (SC) & 84.2 & 58.6 & \textbf{70.5} & \textbf{68.8} & 47.3 & 64.8 & \textbf{84.2} & 68.3 & \\
\midrule
\multirow{3}{*}{Qwen3-8B $\times$12} & RC & \textbf{81.6} & \textbf{49.5} & \textbf{64.7} & \textbf{60.4} & \cellcolor{hl}\textbf{56.6} & 63.2 & \textbf{79.3} & \textbf{65.3} & \\
 & MV (SC) & 80.5 & 48.5 & 64.2 & 60.0 & 56.2 & \textbf{64.0} & 78.3 & 64.9 & \\
\midrule
\multirow{3}{*}{OLMo-3-7B $\times$12} & RC & 76.5 & \textbf{40.4} & \textbf{47.4} & 61.2 & \textbf{48.0} & \textbf{58.0} & \textbf{70.0} & \textbf{58.1} & \\
 & MV (SC) & \textbf{76.8} & \textbf{40.4} & 45.8 & \textbf{63.2} & 46.9 & 54.4 & \textbf{70.0} & 57.6 & \\
\bottomrule
\end{tabular}%
}
\caption{Accuracy (\%) measured in seven benchmarks. \textbf{(a)} Mixed model 12 trace \textsc{RC}, majority vote (MV), Mixture-of-Agents (4 models $\times$ 3 layers) and ReConcile (4 models $\times$ 3 rounds) \textbf{(b)} Single-model RC on 12 traces. Wtd. Avg: weighted mean over all questions. Tok/q: total tokens per question.   \textbf{Bold}/\underline{underline}: best/second-best per column in (a).\textbf{Bold}: best per column within each pool in (b). Shading: best \textsc{RC} pool per column in (b). \textbf{Takeaway: \textsc{RC} outperforms MV in single model pools. Heterogenous model RC outperforms homogenous model RC.} }
\vspace{-10pt}
\label{tab:accuracy}
\end{table*}

\paragraph{Baselines.} Majority voting (MV) over the final answer using the same 12 traces
isolates the benefits of RC's reasoning structure aggregation. Mixture-of-Agents
\citep{wang2024mixtureofagentsenhanceslargelanguage} and ReConcile
\citep{chen2024reconcileroundtableconferenceimproves} are multi-agent debate baselines. We use the same number of generations using the same model to match the inference budget. We also report single-model results for the baseline methods to show how the reasoning diversity of different models leads to better performance. Ties are broken using a random draw. 

\paragraph{Evaluation.} We report the mean final accuracy over three independent runs with different seeds. To compare different methods, we apply exact McNemar tests, paired
bootstrap intervals, and a hierarchical Bayesian model over datasets and seeds (Appendix~\ref{app:full}). We further evaluate the correctness and utility of the consensus graph with a blinded
LLM judge and a user study (\S\ref{sec:judge}).

\vspace{3pt}

\noindent Reporoducibility information in Appendix \ref{app:repro}

%% file: section/05_results.tex
\section{Results}
\label{sec:results}

We compare accuracy with majority voting and debate baselines (\S\ref{sec:accuracy}), analyze
where the structural ensemble helps (\S\ref{sec:structure}), ablate our
design choices (\S\ref{sec:design}), evaluate the consensus with an LLM
judge and users (\S\ref{sec:judge}) and validate the pipeline (\S\ref{pip}).

\subsection{Accuracy Results}
\label{sec:accuracy}

\paragraph{Overall, RC performs at par with MV and debate in accuracy while returning an
auditable graph.} RC reaches 69.4\%, against 69.0\%
for majority voting, 69.5\% for ReConcile and 70.3\% for Mixture-of-Agents
(Table~\ref{tab:accuracy}(a)). Because enforcing reasoning structures often degrades performance \citep{jerge_pitfalls_2025}, RC offers a significant advantage by providing principled explanations of the reasoning process without compromising performance. Cost analysis in Appendix~\ref{app:cost}.

\paragraph{RC performance varies across task types.} Table \ref{tab:fixbreak} depicts a per-dataset count of cases where RC beat a baseline and vice versa. RC's performance varies across three task categories. First, on
\emph{expert knowledge application tasks} (GPQA-D, SARA, Medbullets), RC significantly beats MV (58 wins/33 losses, $p{=}0.011$). RC on average loses to debate on these tasks, but its loss is not statistically significant (123 wins/152 losses v. MoA: $p{=}0.091$, 135 wins/ 146 losses v. Reconcile: $p{=}0.51$). Second, on \emph{narrative evidence weighing tasks} (MuSR), \textsc{RC} is at par with MV. Here, RC significantly beats both debate methods on MuSR-OP ($p{<}0.001$), possibly due to debating
 agents converging on a shared error
\citep{wynn2025talkisntcheapunderstanding}. Third, on \emph{formal entailment} (FOLIO), \textsc{RC} trails all baselines, but none of the losses are significant.

\begin{figure*}[t]
  \centering
  \includegraphics[width=\textwidth]{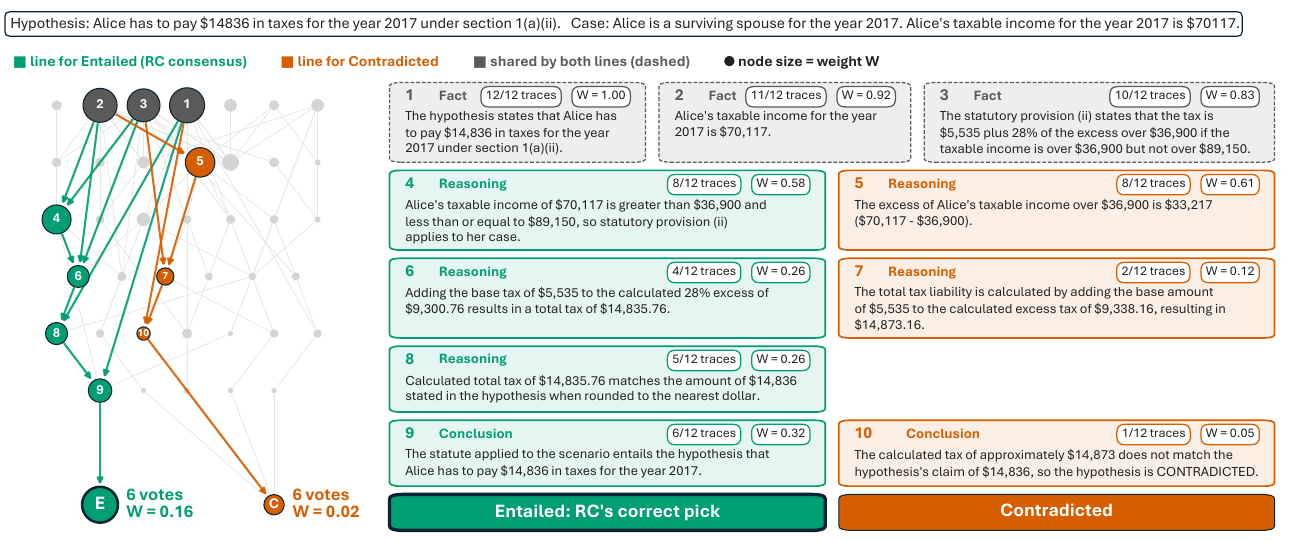}
  \caption{A tied vote that \textsc{RC} resolves (SARA). The 12 traces split 6--6 between \emph{Entailed} (correct) and \emph{Contradicted}, and majority voting's tie-break picks \emph{Contradicted}. \textbf{Left:} the merged graph of the question (41 steps from 12 traces; node size grows with the number of traces stating the step), with the strongest subgraph behind each answer coloured. \textbf{Right:} the steps of both subgraphs, with the number of traces stating each. Both subgraphs start from the same facts (nodes 1--3, stated by 10--12 traces). The consensus subgraph applies the statute's tax bracket and computes tax to conclude \emph{entailment}. The \emph{Contradicted} subgraph miscomputes 28\% of \$33{,}217. This wrong step is corroborated by only two traces. Stronger reasoning support makes \textsc{RC} prefer correct answer \emph{Entailed}. \textbf{Takeaway: RC shows users where losing traces went wrong and what winning traces agreed on.}}
\vspace{-12pt}
  \label{fig:example}
\end{figure*}

\paragraph{Heterogeneous ensembles beat homogeneous ones and
\textsc{RC} helps in both.} (Figure ~\ref{tab:accuracy}b) compares RC using a \textit{single} model vs. \textit{mixed} models with 12 traces. Single-model \textsc{RC} has a consistent advantage over single-model MV (Self-consistency \citep{wang2023selfconsistencyimproveschainthought}). Pooled over the four models and all datasets, single-model \textsc{RC} is right on 101 questions where MV is wrong vs. 72 vice versa ($p{=}0.03$). As different models exhibit strength on different tasks (shaded blue), combining models from different families improves performance by preserving strengths and buffering weaknesses. Mixed-model RC beats single-model RC for all models when pooled over the 7 datasets, with statistically significant gains ($p{<}0.01$) on both Qwen-3-8B and OLMo-3-7B.
We quantitatively analyze that multi-model traces share fewer steps than single-model traces (28.5\% vs.\ 36.9\%), indicating true reasoning diversity beyond final answers.

\begin{table}[t]
\centering
\scriptsize
\setlength{\tabcolsep}{3pt}
\begin{tabular}{@{}lccc@{}}
\toprule
\textbf{Dataset} & \textbf{RC vs.\ MV} & \textbf{RC vs.\ MoA} & \textbf{RC vs.\ ReConcile} \\
\midrule
SARA & 17/10 & \textbf{47/27}$^{*}$ & 47/31 \\
GPQA-D & 25/18 & \textit{45/73}$^{*}$ & 53/64 \\
MedBullets & \textbf{16/5}$^{*}$ & \textit{31/52}$^{*}$ & 35/52 \\
MuSR-MM & 21/22 & \textit{30/72}$^{***}$ & \textit{36/58}$^{*}$ \\
MuSR-OP & 15/12 & \textbf{90/50}$^{***}$ & \textbf{84/39}$^{***}$ \\
MuSR-TA & 9/11 & 44/46 & 39/48 \\
FOLIO & 7/13 & 24/35 & 23/31 \\
\midrule
\textbf{Pooled} & 110/91 & 311/355 & 317/323 \\
\bottomrule
\end{tabular}
\caption{\textbf{Paired comparison analysis:}  \textsc{RC} with each method over (question, seed) pairs. $b/c$: questions only \textsc{RC} / only the other method answers correctly. Exact McNemar test: $^{*}p{<}.05$, $^{**}p{<}.01$, $^{***}p{<}.001$; \textbf{bold}: \textsc{RC} better, \textit{italic}: \textsc{RC} worse. }
\vspace{-12pt}
\label{tab:fixbreak}
\end{table}

\subsection{When Does Structure Help?}
\label{sec:structure}

\paragraph{\textsc{RC} changes the answer when the vote is close, by calculating support through corroboration.} Figure~\ref{fig:example} shows a tie that \textsc{RC} resolves correctly. Of the 4{,}857 question--seed pairs, RC
makes 110 fixes and
91 breaks to MV (Table \ref{tab:fixbreak} ), giving posterior $P$(RC is better than MV) 0.84 on all data and 0.97 on \emph{expert knowledge application tasks}. These changed answers concentrate where the vote is close. On the 5.3\%
of questions where the vote is tied (not necessarily a two-way tie), \textsc{RC} is right 49.4\% of the time,
against 40.9\% random tie-break ($+8.6 pts.[+0.4, +16.9]$). This result shows how a minority answer can win in RC in
close votes when its steps are corroborated more consistently. This reversal happens when the traces of the minority answer share more of their steps with each other, compared to those of the majority answer. Corroboration is most useful on expert-knowledge questions, where correct traces tend to state the same facts and rules.


\paragraph{RC is suboptimal when diverse plausible explanations compete.} RC does not outperform MV in the narrative MuSR dataset. 
In this dataset, correct traces tend to use distinct approaches to reach the answer, as witnessed by the low cluster overlap between traces (18--26\%, compared to 35\%
on SARA). On FOLIO, \textsc{RC} demonstrated weakness when the answer is \emph{uncertain} (neither entailment nor contradiction), with 9 regressions and 1 fix. This can be attributed to the fact that the uncertain label can be deduced from many possible explanations as long as they deny \textit{True} and \textit{False}, leading to low overlap between traces. Still, the structural representations of RC allow users to inspect different reasoning subgraphs. Further failure diagnosis results can be found in
Appendix~\ref{app:diagnosis}.

\subsection{Ablation studies}

\label{sec:design}

\paragraph{DAG extractor model.} Extracting the DAGs with
Qwen3-8B instead of Qwen3.8-27B lowers \textsc{RC} from 69.9\% to 68.3\%, below MV (68.9\%). This indicates that RC requires the DAG extractor to be accurate to a certain level; otherwise, the DAG extraction error accumulates and can harm the end-to-end performance.

\paragraph{Merging strategies.} Our approach clusters steps by embedding similarity and
verifies each cluster with an LLM call. We test two variations: Pairwise merging, where the clustering is done by asking an LLM if two steps are equivalent, and Cluster-only, where we skip the verification stage. Pairwise merging (asking the LLM about whether two steps should merge) requires
1{,}247 inference calls per task, $\sim 75\times$ more than the main approach (16.6 calls), with statistically insignificant difference in accuracy. On the other hand, cluster-only performance requires less inference, but the accuracy drops to 67.5\% ($p{=}0.007$).

\paragraph{Number of traces $|\mathcal{T}|$.} We ran an ablation with 4, 8, and 12 traces per question. On 4
traces, \textsc{RC} matches MV (67.5\% vs.\ 67.4\%). With 8 and
12 traces, RC leads MV (68.9\% vs.\ 68.3\% and 69.9\% vs.\ 68.9\%).
This indicates that RC scales better than MV by leveraging the additional step information and corroboration not used in MV.

 \vspace{6pt}

\noindent Detailed results on ablation results in Appendix \ref{ablation}.

\subsection{Evaluating Explanation Quality}
\label{sec:judge}

\begin{table}[t] 
\centering
\scriptsize
\setlength{\tabcolsep}{4pt}
\begin{tabular}{@{}lrcc@{}}
\toprule
\textbf{Dataset} & \textbf{$n$} & \textbf{Raw} & \textbf{Length debiased} \\
\midrule
SARA       & 272 & 0.58 & \textbf{0.58} {\tiny[0.51, 0.64]} \\
GPQA-D     & 198 & 0.61 & 0.56 {\tiny[0.47, 0.63]} \\
MedBullets & 190 & 0.69 & \textbf{0.69} {\tiny[0.61, 0.76]} \\
MuSR-MM    & 250 & 0.54 & \textbf{0.62} {\tiny[0.55, 0.69]} \\
MuSR-OP    & 256 & 0.53 & \textbf{0.58} {\tiny[0.50, 0.65]} \\
MuSR-TA    & 250 & 0.52 & \textbf{0.61} {\tiny[0.53, 0.67]} \\
FOLIO      & 203 & 0.56 & \textbf{0.59} {\tiny[0.51, 0.66]} \\
\midrule
\textbf{Pooled} & 1{,}619 & 0.57 & \textbf{0.61} {\tiny[0.58, 0.64]} \\
\bottomrule
\end{tabular}
\caption{LLM-judge preference for the consensus subgraph (Claude Opus 5.5, all questions, each pair judged in both orders) over a random subgraph supporting the same answer. \textbf{Bold}: preference > 0.5 with 95\% confidence, after debiasing the length effect. \textbf{Takeaway: LLM judge prefers Consensus subgraph.}}
\vspace{-12pt}
\label{tab:judge}
\end{table}

\paragraph{LLM judge evaluation.} High accuracy does not imply better reasoning explanation quality \citep{lee_evaluating_2025}. For each consensus answer (total 1{,}619), we
ask Claude Opus 5.5 (a model not used for trace generation and DAG extraction), to compare the consensus graph (most attested explanation) with a random subgraph leading to the same answer.
The judge sees the problem and both explanations, not knowing whether the answer is correct or which is the consensus graph. It compares two explanations in three aspects: grounding, soundness, and gaps, and produces a final verdict between options A and B. Since LLM judges tend to have a length bias, favoring
longer explanations \cite{hu2025explaininglengthbiasllmbased}, we also report
preferences after debiasing the length effect, estimated with a hierarchical model that includes
length and order as covariates (Appendix~\ref{app:judge}).
At equal length, the judge prefers the consensus with probability 0.61
$[0.58, 0.64]$ (Table~\ref{tab:judge}). This preference holds on every
benchmark, most clearly on MedBullets (0.69). \textit{\textbf{Takeaway: The LLM judge prefers the consensus subgraph for the same answer}.}

\paragraph{Human evaluation.} Next, we inspect whether the reasoning consensus graphs are more inspectable to humans than raw traces and debate results. 18 student volunteers each evaluated 6 explanations drawn from MuSR, SARA, and
MedBullets in three formats---MV with 12 raw reasoning traces, Reconcile's debate transcript, and RC consensus graph along with the strongest graph supporting the minority answer. Each participant
saw exactly 3 samples where the LLM majority voting is correct and 3 incorrect samples (one of each per format).
Participants identified the central point of
reasoning for the majority answer and the step where differing solutions diverged equally well across formats
(Accuracy: graph 82\%, vote 82\%, debate 79\%), but the graph required the least time and effort:
a median of 3.4 active minutes per question (vote 5.1, debate 4.4) and
a median effort of 2.5 on a 1--5 scale (vote 4, debate 3). In the post-experiment survey, 14/18
participants ranked RC graphs to be the easiest, and 14/18 preferred the RC format for future use. This demonstrates how RC graphs considerably reduce the time and effort required to understand multiple reasoning traces at once. Details in Appendix \ref{app:human-study}. \textit{\textbf{Takeaway: Users of RC require lesser time and effort to understand the main point of reasoning for an answer and divergent reasoning for alternative answers.}}

\subsection{Pipeline Validation}
\label{pip}

We analyze the correctness of the components of our pipeline, DAG extraction and merging clusters.

\paragraph{Automated validation.} \textit{\textbf{DAG extraction validity:}} LLM-based audit using Claude Opus 5.5 evaluated 95.3\% of the extracted DAGs as valid. \textit{\textbf{Merge validity:}} We asked the LLM judge whether the clusters with similarity below the threshold $\tau{=}0.55$ should be merged. The judge would have merged only 1.7\% of these, showing the high recall for clustering. Deailed pipeline validation in Appendix \ref{analyses} 

\paragraph{Manual validation.} The same annotators evaluated the extraction validity and merge validity. \textit{\textbf{DAG extraction validity:}} on 90 extracted reasoning
steps (5 per person, 3 raters each), annotators judged 64\% as fully
grounded in the source text, 32\% as partly grounded, and 3\% as
unsupported, confirming that the extractor paraphrases and occasionally
composes across sentences but rarely fabricates.
\textit{\textbf{Merge validity:}} on 270 step pairs (15 per person,
3 raters each), annotators agreed with the pipeline's merge decisions for
83\% of merged pairs (judged ``same'') and 98\% of non-merged pairs
(judged ``different''), supporting that the automatic merge is highly accurate. More details and instructions are in Appendix \ref{app:human-study}.

\subsection{Qualitative Analysis}
\label{sec:qualitative}

\paragraph{The graph shows \textit{where} traces diverge.} In
Figure~\ref{fig:example}, both answers start from the same facts, and the
losing answer can be traced back to one miscomputed step stated by two traces. A reader
who wants to check the answer can verify that step instead of reading all twelve traces.

\paragraph{RC graph reveals correct minority reasoning that MV discards.} One of the SARA
questions (Appendix \ref{app:examples}) asks whether Alice is related to Bob under a statute covering
``a son or daughter of a brother or sister of the taxpayer'' (answer: True). 8 of 12
traces answer \emph{False}, incorrectly assuming that Alice is the taxpayer and concluding that Bob is Alice's uncle, not niece.
Only the remaining four
traces correctly reason that Alice is the daughter of Bob's brother and conclude
\emph{True}. Majority voting reports only the 8--4 split, while RC
shows the minority reasoning path and the one step on which the answers part. We see that in 80\% of the cases where \textsc{RC} and MV are both wrong, at least one trace reaches the gold answer, so its reasoning is preserved in the ensemble graph.

\paragraph{One answer, several reasoning approaches.} The merged graph can hold more than one reasoning approach
 for the same answer. The subgraphs we compare the consensus subgraph against in
Section~\ref{sec:judge} reach the same answer, yet share
only 29\% of their merged steps with the consensus (median). In a GPQA-D
question, four traces compute a measurement probability through an inner
product and four through a ratio of projection norms, and both routes remain
in the graph for a reader to check 
\vspace{3pt}

\noindent(More examples can be found in Appendix~\ref{app:examples}).

\section{Discussion}

 RC matches voting in accuracy, gains most when correct traces share steps, and produces a consensus graph that judges and users prefer. 

\paragraph{Corroboration helps when correct reasoners share steps.}
RC changes the voted answer mainly when the vote is close, and gains most on expert-knowledge questions, where correct traces state the same facts and rules.
In cases where correct traces take different aproaches (MuSR), there is less corroboration to measure.
There, RC's value lies in exposing alternate approaches than in accuracy.

\paragraph{Independence keeps disagreement visible.}
No model sees another's reasoning, so errors cannot spread through discussion.
Minority arguments also survive in the merged graph and users can see where the reasoning for different answers diverges.

\section{Conclusion}

We presented \textsc{Reasoning Consensus} (RC), which merges independently generated reasoning traces from diverse models into a single weighted graph and reveals the most-supported approach.
RC matches majority voting and multi-agent debate in accuracy. Its most distinctive gain comes on tied votes, where it raises accuracy by 8.6 points over random tie-breaking.
Because each step's weight reflects how many independent traces agree on it, RC also provides step-level confidence without extra model calls.
The consensus reasoning subgraph it returns is preferred by a blinded judge over other subgraphs that reach the same answer.
Ablations of RC's components, qualitative analysis of the merged graphs, and a user study further support these findings.
RC therefore returns more than an answer. It shows the different approaches and answers produced by varied models, together with their step-wise support, so users can make an informed choice.


%% file: section/08_limitations.tex
\section*{Limitations}

\paragraph{Support is corroboration, not verification.} A high attestation count
means that several independent traces stated a step, not that the step is
correct. An error common to several models gains weight just as genuine
agreement does. \textsc{RC} mitigates this by never discarding the alternatives.
Competing conclusions are kept with their support, so a reader sees when traces
disagreed instead of receiving a single confident answer. The graph is
therefore inspectable rather than certified. Its value is that a reader can
inspect each step and reject it, which a prose rationale does not allow.
A natural extension is a verification pass over the highest-weight steps:
each step could be checked against the problem statement and its parent steps,
either by a verifier model or, where possible, by symbolic execution of
arithmetic and code. Verified support could then be weighted separately from
raw attestation.

\paragraph{Cost and dependence on the extractor.} \textsc{RC} uses 4.1 times the
tokens of majority voting (65.1k vs.\ 15.8k per question), half of them for
extracting the DAGs. With a weaker extractor (Qwen3-8B instead of Qwen3.8-27B),
it falls below voting (68.3 vs.\ 68.9), so it needs a strong
extraction model.
Since extraction is a narrow, well-specified task, a small model fine-tuned on
DAGs produced by the strong extractor may recover most of its quality at a
fraction of the cost. Prompting the reasoning models to emit steps in a
structured format would also remove much of the extraction work.

\section*{AI Involvement Disclosure}
AI assistance was limited to language polishing for grammar and readability, polishing of LLM prompts, and improving code readability. The conception, methodology, experiments, analyses, and interpretations were conducted fully by the authors.

\section*{Acknowledgement}
This work used NCSA Delta (NSF OAC 2005572) and DeltaAI (NSF OAC 2320345) through ACCESS allocation CIS251114, supported by NSF grants \#2138259, \#2138286, \#2138307, \#2137603, and \#2138296. This work was also supported in part by a fellowship from Amazon.

\bibliography{custom}

%% file: section/A1_prompts.tex
\newpage

\appendix

\section{Prompts}
\label{app:prompts}

\subsection{Trace generation}
\label{app:prompts:gen}

\begin{prompt}{System prompt (all benchmarks).}
You are a careful reasoner. For every question, you must:
1. Think step by step inside a single <reasoning>...</reasoning> block.
2. Put ONLY the final answer (no explanation) inside a <answer>...</answer> tag.
Do not write anything outside these two tags. Do not abbreviate your reasoning.
\end{prompt}

\begin{prompt}{SARA.}
You are reasoning about US federal tax law.

STATUTE:
{statute}

CASE SCENARIO:
{case}

HYPOTHESIS:
{hypothesis}

TASK: Decide whether the statute, applied to the case scenario, ENTAILS or CONTRADICTS the hypothesis. Walk through:
  (a) which statutory provisions apply to the case,
  (b) what facts in the scenario trigger or fail to trigger those provisions,
  (c) what the provisions imply about the hypothesis.

Output exactly:
<reasoning>your step-by-step legal analysis</reasoning>
<answer>LABEL</answer>   where LABEL is exactly one of ENTAILED or CONTRADICTED
\end{prompt}

\begin{prompt}{GPQA Diamond.}
You are answering a graduate-level science question. Exactly one option is correct.

QUESTION:
{question}

OPTIONS:
  A. {option A}
  B. {option B}
  C. {option C}
  D. {option D}

TASK: Reason carefully through the underlying science. Walk through:
  (a) what principles or equations the question is testing,
  (b) how to apply them to the specifics of this problem,
  (c) why one option follows and the others do not.

Output exactly:
<reasoning>your step-by-step scientific reasoning</reasoning>
<answer>LETTER</answer>   where LETTER is exactly one of A, B, C or D (a single capital letter only)
\end{prompt}

\begin{prompt}{MedBullets.}
You are answering a clinical medicine question (USMLE Step 2/3 style). Exactly one option is correct.

QUESTION:
{question}

OPTIONS:
  A. {option A}
  B. {option B}
  C. {option C}
  D. {option D}

TASK: Reason carefully through the clinical case. Walk through:
  (a) the key findings in the case (history, examination, investigations),
  (b) what diagnosis or mechanism they point to and how that bears on what is asked,
  (c) why one option follows and the others do not.

Output exactly:
<reasoning>your step-by-step clinical reasoning</reasoning>
<answer>LETTER</answer>   where LETTER is exactly one of A, B, C or D (a single capital letter only)
\end{prompt}

\begin{prompt}{MuSR (all three domains).}
You are reasoning about a narrative scenario. Read the narrative carefully, then choose the single correct answer.

NARRATIVE:
{narrative}

QUESTION:
{question}

OPTIONS:
  A. {option A}
  B. {option B}
  ...

TASK: Reason carefully through the evidence in the narrative. Walk through:
  (a) what facts the narrative establishes,
  (b) how those facts bear on each candidate option,
  (c) why one option is best supported and the others are not.

Output exactly:
<reasoning>your step-by-step reasoning over the narrative</reasoning>
<answer>LETTER</answer>   where LETTER is exactly one of {A, B, ...} (a single capital letter only)
\end{prompt}

\begin{prompt}{FOLIO.}
You are reasoning about first-order logical entailment in natural language.

PREMISES:
{premises}

CONCLUSION:
{conclusion}

TASK: Decide whether the conclusion is logically entailed by the premises. Use these labels:
  TRUE      -- the conclusion follows necessarily from the premises
  FALSE     -- the negation of the conclusion follows from the premises
  UNCERTAIN -- neither the conclusion nor its negation follows

Walk through:
  (a) what each premise asserts,
  (b) what chain of inferences (if any) the conclusion would require,
  (c) whether the premises are sufficient to establish or refute it.

Output exactly:
<reasoning>your step-by-step logical analysis</reasoning>
<answer>LABEL</answer>   where LABEL is exactly one of TRUE, FALSE or UNCERTAIN
\end{prompt}

\subsection{DAG extraction}
\label{app:prompts:extract}

\begin{prompt}{}
You are a structured-reasoning extractor. Given a question, a reasoning trace and its final answer, you produce a directed acyclic graph (DAG) in JSON that captures the trace's logical structure. Output only the JSON object.

TASK: Extract a Type-B reasoning DAG from the reasoning trace below.

NODE TYPES (use exactly these strings):
  - "Planning":   meta-reasoning, problem decomposition, deciding what to check
  - "Fact":       a factual statement drawn from the question context (statute, scenario, narrative, premises,
                  options); its text should restate the relevant span of the question
  - "Reasoning":  an intermediate inferential step linking facts to the outcome
  - "Conclusion": the final determination

REQUIRED: exactly ONE node of type "Conclusion", whose text expresses the final answer given below (in the
trace's own words, but the same verdict). All other nodes must transitively support this Conclusion.

Each node has:
  - "id":      unique short id ("p1", "f1", "r1", "c1", ...)
  - "type":    one of the four types above
  - "text":    one self-contained sentence (resolve pronouns and references like "option B" or "the
               hypothesis" using the question context)
  - "support": list of bundles. A bundle is a list of node ids that JOINTLY (AND) justify this node;
               different bundles are alternative, independent (OR) justifications.
               states a counter-consideration. Do not invent defeaters.
  - "rejects": the ANSWER LABELS (listed below the question) that this step explicitly argues AGAINST, e.g.
               "Option A is wrong because ..." -> ["A"]; "the conclusion cannot be shown to be true" -> ["TRUE"].
               [] for most steps. Only what the trace itself says; never the final answer's own label.

RULES:
  - List nodes premises first: all Fact nodes first, then Planning and Reasoning nodes, the Conclusion last.
  - A bundle may only reference ids of nodes listed EARLIER (this keeps the graph acyclic).
  - Facts may have empty support (atomic premises from the question).
  - Planning, Reasoning and Conclusion nodes MUST have at least one support bundle.
  - A bundle names only the few steps that directly justify the node (at most 10 ids); at most 4 bundles per node.
  - Do NOT add information the trace does not contain. One node per distinct claim; at most 60 nodes.

----------------------------------------
QUESTION CONTEXT:
{question context}

ANSWER LABELS: {labels}

FINAL ANSWER (the Conclusion node's verdict):
{answer}

REASONING TRACE:
{trace}
----------------------------------------

Output only the JSON object {"nodes": [...]}.
\end{prompt}

\subsection{Merge verification}
\label{app:prompts:merge}

\begin{prompt}{}
You are a strict judge of semantic equivalence between statements extracted from reasoning traces. You group statements that make the SAME claim. Output only JSON.

QUESTION CONTEXT:
{question context}

----------------------------------------
The numbered statements below are "{node type}" steps taken from different reasoning traces about this question. Group together the statements that state the SAME claim (same entities, same assertion, same polarity), regardless of wording.
Rules:
  - Lexical overlap is not enough. Different entities, numbers, provisions, options or inferences are DIFFERENT claims.
  - A general rule and its application to a specific case are DIFFERENT claims.
  - A claim and its negation are DIFFERENT claims.
  - Every statement appears in exactly one group; a statement that matches no other is a group of its own.

STATEMENTS:
1. {statement}
2. {statement}
...

Output JSON: {"groups": [[1, 3], [2], ...]}
\end{prompt}

\subsection{Baselines}
\label{app:prompts:baselines}

\begin{prompt}{Mixture-of-Agents (the aggregate-and-synthesize prompt of \citealp{wang2024mixtureofagentsenhanceslargelanguage}, adapted to our format)}
You have been provided with a set of responses from various models to the question below. Your task is to synthesize these responses into a single, high-quality answer. It is crucial to critically evaluate the information provided in these responses, recognizing that some of it may be biased or incorrect. Your response should not simply replicate the given answers but should offer a refined, accurate reasoning and answer. You may agree with one of the responses or reach a better answer from their reasoning.

=== ORIGINAL TASK ===
{generation prompt}

=== RESPONSES FROM MODELS ===
--- Response 1 ---
Reasoning:
{reasoning}
Final answer: {answer}
...

=== YOUR TURN ===
Now give your own carefully reasoned answer in exactly the format required by the original task: <reasoning>...</reasoning> then <answer>...</answer>.
\end{prompt}

\begin{prompt}{ReConcile, system prompt.}
You are a careful reasoner. For every question, you must:
1. Think step by step inside a single <reasoning>...</reasoning> block.
2. Put ONLY the final answer (no explanation) inside a <answer>...</answer> tag.
3. Give your confidence that your answer is correct, a number between 0.0 and 1.0, inside a <confidence>...</confidence> tag.
Do not write anything outside these three tags. Do not abbreviate your reasoning.
\end{prompt}

\begin{prompt}{ReConcile, round 1}
Also evaluate your confidence level (between 0.0 and 1.0) to indicate the possibility of your answer being right, and put it after the answer as <confidence>0.0-1.0</confidence>, e.g. <confidence>0.8</confidence>.
\end{prompt}

\begin{prompt}{ReConcile, rounds 2--3.}
You and other agents answered the question below. Carefully review the solutions from the other agents as additional information, and provide your own answer and step-by-step reasoning to the question. Clearly state which points of view you agree or disagree with and why. You may keep your previous answer if you still believe it is correct, or change it if another agent's reasoning is more convincing.

=== ORIGINAL TASK ===
{generation prompt}

=== YOUR PREVIOUS SOLUTION ===
Answer: {answer} (confidence: {confidence})
Reasoning:
{reasoning}
Final answer: {answer}

=== OTHER AGENTS' SOLUTIONS ===
--- Agent 1 ---
Answer: {answer} (confidence: {confidence})
...

=== YOUR TURN ===
Now give your reconsidered answer in exactly the format required by the original task: <reasoning>...</reasoning> then <answer>...</answer>, followed by <confidence>0.0-1.0</confidence>.
\end{prompt}

\subsection{LLM judge}
\label{app:prompts:judge}

\begin{prompt}{}
You are an expert evaluator of written reasoning. You will be shown a problem and candidate explanations of an answer to it. Each explanation is a numbered list of steps; each step is labelled with its role (Planning, Fact, Reasoning, Conclusion, Answer).

Judge each explanation as a justification of its own final answer, using only these criteria:
1. Grounding: the facts and premises it uses are stated in the problem or follow correctly from it; it does not invent, distort or misquote information.
2. Soundness: every inference follows validly from the steps it builds on.
3. No gaps or irrelevant leaps: the steps connect the evidence to the final answer without unexplained jumps, and without steps that do not bear on the answer.

Length is NOT a criterion. An explanation is not better because it has more steps or more words, and not worse because it is concise: judge only whether the steps it gives are grounded, sound and sufficient. The order in which explanations are shown is arbitrary and carries no information.

PROBLEM:
{question context}

Both explanations below conclude the same final answer: {answer}. Decide which one justifies it better.

=== EXPLANATION A ===
{explanation}

=== EXPLANATION B ===
{explanation}

Which explanation is the better justification of the answer? Reply with a single letter: A or B.
\end{prompt}

\section{Method Details}
\label{app:method}

\subsection{Extraction}

We validate node types and acyclicity, applying
deterministic repairs if needed. Irreparable traces are only used for
final-answer vote. These deterministic fixes are applied in order:
\begin{enumerate}
\item drop malformed nodes and duplicate ids, and map unknown node types to the
  closest type.
\item drop self-references from premises, deduplicate premises,
  and drop empty ones.
\item keep only one Conclusion by merging repeat.
\item attach every unsupported non-Fact step to the nearest earlier
  Fact step. Reasoning step restating the question is
  retyped as a Fact.
\item break any remaining cycles (repeat steps 4 and 5 till neither applies)
\end{enumerate}

\subsection{Merging}
\label{app:method:merge}

We embed all steps using the all-mpnet-base-v2 sentence embedding model, and cluster them separately per step type (Planning, Fact, Reasoning, and Conclusion). Then, each cluster is split into
sub-clusters $S_j$ so that steps from the same traces are placed into different clusters. Specifically, we process steps in ascending order of distance from the cluster mean, assigning each to the most similar sub-cluster that contains no other step from the same trace, or creating a new sub-cluster if none exists.
An LLM then partitions the resulting sub-clusters into groups containing semantically equivalent steps. 
Answer nodes always merge by their extracted final answer, so that each unique final answer will map to exactly one node.

\subsection{Scoring}
\label{app:method:score}

\paragraph{Attestation.} For a Reasoning, Planning, or Conclusion cluster,
$\mathcal{T}(c)$ is the set of traces that attest at least one of its
premise sets. For a Fact cluster, every trace that states it attests it, and
$\Phi(c) = 1$ if at least one trace states it without support. Answer
node $a_y$ is attested by every trace predicting $y$.

\paragraph{Answers without structure.} An answer predicted only by traces
with answer-only DAGs has no support bundle to score. We impute its support
signal as the mean $\Phi$ of the question's other answers, i.e.\ the mean of
$W(a_{y'}) / \alpha(a_{y'})$, so that $W(a_y) = \alpha(a_y)$ times this
mean. If no answer of the question has structure, every $\Phi$ is 1 and
\textsc{RC} reduces to majority voting.

\paragraph{Well-foundedness.} $W(c)$ depends on the scores of the $c$'s incoming hyperedges, which are obtained from scores of other clusters. This recursion is
well-founded: the ensemble DAHG is acyclic (validated at extraction and restored
after merging), so its nodes admit a topological order in which every bundle
member precedes the node it supports or defeats. A single pass in that order
computes every score from already final values, and exactly one assignment
satisfies Eqs.~\ref{eq:phi}--\ref{eq:score}.

\paragraph{Ties.} The answer is $\arg\max_y W(a_y)$. Ties are broken by the
number of traces predicting each answer. Remaining ties by a fixed random
draw per question that all methods share, not by label order.

\section{Pipeline Validation Analysis}
\label{analyses}
\subsection{Diversity of correct traces}
\label{app:diversity}

Mixing models widens coverage and diversity of justifications.
On seed 1234, a mixed pool (3 traces $\times$ 4 models) contains a correct
trace for 93.6\% of questions vs.\ 83.4--87.0\% for 12 traces from any single
model, highest on all 7 datasets, despite averaging fewer correct traces than
Gemma or Phi-4 alone (Table~\ref{tab:diversity}). Correct traces from different models share fewer steps than correct
traces from the same model (28.5\% vs.\ 36.9\%).

\begin{table}[htbp]
\centering
\small
\setlength{\tabcolsep}{2.5pt}
\begin{tabular}{@{}lrrrrrr@{}}
\toprule
 & $n$ & \textbf{Mixed} & Gemma & Qwen & OLMo & Phi \\
\midrule
FOLIO      & 203  & \textbf{98.0} & 93.1 & 96.1 & 96.1 & 92.6 \\
GPQA       & 198  & \textbf{94.4} & 85.9 & 77.3 & 84.8 & 87.4 \\
MedBullets & 190  & \textbf{91.6} & 83.7 & 78.9 & 80.0 & 89.5 \\
MuSR-MM    & 250  & \textbf{97.2} & 84.4 & 76.8 & 86.8 & 89.6 \\
MuSR-OP    & 256  & \textbf{83.6} & 64.5 & 77.7 & 74.6 & 71.5 \\
MuSR-TA    & 250  & \textbf{92.4} & 86.0 & 82.4 & 88.4 & 84.4 \\
SARA       & 272  & \textbf{98.2} & 96.0 & 94.1 & 97.1 & 94.1 \\
\midrule
All        & 1619 & \textbf{93.6} & 84.6 & 83.4 & 87.0 & 86.8 \\

\bottomrule
\end{tabular}
\caption{Share of questions (\%) whose 12-trace pool contains at least one
correct trace: mixed (3 traces $\times$ 4 models) vs.\ 12 traces from one
model. Seed 1234. }
\label{tab:diversity}
\end{table}

\subsection{DAG Extraction Validity}
\label{app:extract:validity}

Every extracted DAG is parsed and checked for permitted node types, valid
premise references, exactly one Conclusion, support for every non-Fact step and
acyclicity. Table~\ref{tab:validity} reports how many DAGs pass as extracted,
how many pass after the repairs , and how
many fall back to an answer-only DAG.

\begin{table}[htbp]
\centering
\small
\setlength{\tabcolsep}{3pt}
\begin{tabular}{@{}lrrrr@{}}
\toprule
\textbf{Generator} & \textbf{Valid} & \textbf{Repaired}
 & \textbf{Fallback} & \textbf{No ans.} \\
\midrule
Gemma-4-E4B & 89.0 & 11.0 & 0.0 & 0.0 \\
Qwen3-8B    & 96.7 &  2.3 & 0.0 & 1.0 \\
OLMo-3-7B   & 96.8 &  2.6 & 0.0 & 0.6 \\
Phi-4       & 98.7 &  1.3 & 0.0 & 0.0 \\
\midrule
All         & 95.3 &  4.3 & 0.0 & 0.4 \\
\bottomrule
\end{tabular}

\caption{Extraction outcomes (\% of traces; 14{,}571 per generator over seven
benchmarks and three seeds). Valid: passes every check as extracted. Repaired:
passes after deterministic repairs. Fallback: kept as an answer-only DAG. No
ans.: the trace gave no parsable answer.}
\label{tab:validity}
\end{table}

\newpage

95.3\% of DAGs pass every check as extracted and a further 4.3\% after repair;
no trace needed the answer-only fallback. Almost all repairs (4\% of traces)
attach a step that the extractor left unsupported. They concentrate on
Gemma-4-E4B, whose traces on MuSR-TA and FOLIO need a repair in 25\% of cases
and on MuSR-MM and MuSR-OP in 10--13\%, against at most 10\% for any other
generator and benchmark. The other generators' raw validity is 96.7--98.7\%.
The 0.4\% of traces without a parsable answer, mostly on GPQA-D and MuSR-TA,
come from generation, not extraction.

\subsection{Merge threshold: missed merges}
\label{app:tau}

We validate our cosine similarity threshold $\tau$ choice post experimentation. We cluster each question at $\tau \in \{0.45, 0.55, 0.65, 0.75\}$ with the same embedder, average-linkage clustering and same-trace split, and mark each pair as kept (same sub-cluster) or discarded at each $\tau$. We ask the merge judge (Qwen3.8-27B) if each pair states the same claim, using the pairwise YES/NO prompt of the merge step. We then check what percentage of the discarded pairs would have been voted yes, to ensure that our threshold was not removing useful pairs. The percentage of discarded pairs by a threshold shows us the LLM compute saved by the threshold.

\begin{table}[htbp]
\centering
\scriptsize
\setlength{\tabcolsep}{2pt}
\begin{tabular}{@{}lcccc@{}}
\toprule
$\tau$ & 0.45 & \textbf{0.55} & 0.65 & 0.75 \\
\midrule
Discarded pairs (\% of all) & 82.1 & 83.1 & 84.6 & 87.2 \\
Judge YES among discarded (\%) & 2.30 & 1.70 & 1.50 & 2.01 \\

\bottomrule
\end{tabular}
\caption{Step pairs discarded by clustering at each merge threshold, judged by the merge judge . }
\label{tab:tau_missed}
\end{table}

$\tau{=}0.55$, clustering discards 83.1\% of step pairs, sparing the
judge those calls, and the judge would have accepted only 1.7\% of them.
Lowering $\tau$ keeps more pairs for the judge (82.1\% discarded at 0.45)
yet misses the most merges (2.3\%), since loose clusters mix unrelated
steps that later split apart. Raising $\tau$ discards more pairs (87.2\%
at 0.75) and starts dropping similar ones (2.0\% missed). $\tau{=}0.55$
and 0.65 miss fewest merges (1.7\% and 1.5\%, in noise of each
other) (Table~\ref{tab:tau_missed}).

\section{Cost}
\label{app:cost}

\begin{table*}[t]
\centering
\small
\setlength{\tabcolsep}{4pt}
\begin{tabular}{@{}lrrrrrrrrr@{}}
\toprule
 & \multicolumn{2}{c}{\textbf{Generation}}
 & \multicolumn{2}{c}{\textbf{Extraction}}
 & \multicolumn{2}{c}{\textbf{Merging}}
 & & & \\
\cmidrule(lr){2-3}\cmidrule(lr){4-5}\cmidrule(lr){6-7}
\textbf{Method} & in & out & in & out & in & out
 & \textbf{Total} & \textbf{$\times$MV} & \textbf{Calls} \\
\midrule
Majority vote (12 traces)
 & 8{,}981 & 6{,}782 & -- & -- & -- & --
 & 15{,}763 & 1.0 & 12.0 \\
\textsc{RC}
 & 8{,}981 & 6{,}782 & 20{,}813 & 11{,}365 & 16{,}570 & 560
 & 65{,}071 & 4.1 & 40.6 \\
Mixture-of-Agents ($4 \times 3$ layers)
 & 25{,}653 & 5{,}941 & -- & -- & -- & --
 & 31{,}594 & 2.0 & 12.0 \\
ReConcile ($4 \times 3$ rounds)
 & 27{,}314 & 6{,}308 & -- & -- & -- & --
 & 33{,}622 & 2.1 & 12.0 \\
\bottomrule
\end{tabular}
\caption{Tokens and LLM calls per question, by stage (mean over all
questions of the seven benchmarks).}
\label{tab:cost}
\end{table*}

Table~\ref{tab:cost} gives tokens per question from the inference logs
(input and output, mean over all questions and seeds). All methods use the
same 12 generations, so they match in traces but not in tokens. \textsc{RC}
uses 4.1$\times$ the tokens of majority voting and about 2$\times$ those of
the debate methods. 
\newpage
Extraction takes half (49\%), merging a quarter (26\%)
and generation the rest (24\%). Its 40.6 calls are 12 generations,
12 extractions and 16.6 verifier calls, one per multi-step cluster. The debate
methods spend their extra tokens on input, as each layer or round re-reads
the previous responses.

\begin{table}[htbp]
\centering
\small
\setlength{\tabcolsep}{3pt}
\begin{tabular}{@{}lrrrr@{}}
\toprule
\textbf{Dataset} & \textbf{MV} & \textbf{RC} & \textbf{MoA} & \textbf{ReC.} \\
\midrule
SARA       &  9.3k & 41.0k & 24.2k & 26.4k \\
GPQA-D     & 19.6k & 71.4k & 40.6k & 42.8k \\
MedBullets & 12.0k & 53.6k & 27.7k & 29.4k \\
MuSR-MM    & 21.4k & 94.8k & 35.9k & 37.7k \\
MuSR-OP    & 18.8k & 72.3k & 30.8k & 32.6k \\
MuSR-TA    & 17.7k & 73.0k & 34.2k & 36.5k \\
FOLIO      & 11.1k & 46.5k & 28.9k & 31.0k \\
\midrule
Avg & 15.8k & 65.1k & 31.6k & 33.6k \\
\bottomrule
\end{tabular}
\caption{Total tokens per question by benchmark.}
\label{tab:cost-ds}
\end{table}

Cost grows with input and trace length (Table~\ref{tab:cost-ds}). MuSR-MM is
the most expensive benchmark for majority voting and \textsc{RC}, and GPQA-D
for the debate methods; \textsc{RC} costs 3.6--4.5$\times$ majority voting on
every benchmark. A single trace costs 0.6--2.8k tokens, depending on model
and benchmark. Most of \textsc{RC}'s overhead is extraction; a smaller
extractor would cut it but would lower accuracy.

  \section{Ablations}
\label{ablation}

\subsection{A smaller extractor}
\label{app:extract:small}

\begin{table}[htbp]
\centering
\small
\setlength{\tabcolsep}{3pt}
\begin{tabular}{@{}lccccc@{}}
\toprule
\textbf{Dataset} & \textbf{27B} & \textbf{8B} & \textbf{MV}
 & \textbf{$\Delta$} & \textbf{fix/brk} \\
\midrule
SARA       & 86.8 & 85.7 & 85.7 & $-1.1$ & 3/6 \\
GPQA-D     & 60.6 & 56.1 & 58.1 & $-4.5$ & 5/14 \\
MedBullets & 70.5 & 66.8 & 66.3 & $-3.7$ & 1/8 \\
MuSR-MM    & 66.8 & 65.2 & 66.0 & $-1.6$ & 7/11 \\
MuSR-OP    & 52.7 & 51.6 & 53.1 & $-1.2$ & 5/8 \\
MuSR-TA    & 67.2 & 66.8 & 66.8 & $-0.4$ & 6/7 \\
FOLIO      & 84.7 & 85.2 & 85.2 & $+0.5$ & 7/6 \\
\midrule
All        & 69.9 & 68.3 & 68.9 & $-1.7$ & 34/60 \\
\bottomrule
\end{tabular}
\caption{Accuracy (\%) of \textsc{RC} with Qwen3.8-27B or Qwen3-8B as extractor
and merge verifier, and of majority voting, on seed 1234. fix/brk: questions
8B variant gets right / wrong where the 27B version does not.}
\label{tab:extractor}
\end{table}

\paragraph{Experiment:} To test how much \textsc{RC} depends on the extraction model, we replace
Qwen3.8-27B with Qwen3-8B, the size of the generators, for both extraction and
merge verification, keeping everything else fixed (seed 1234, all 1{,}619
questions).

\paragraph{Results:} With the smaller model, \textsc{RC} loses 1.7 points (34 fixes vs.\ 60 breaks,
McNemar $p = 0.01$) and falls below majority voting (68.3\% vs.\ 68.9\%;
33/42 against voting, $p = 0.36$). The loss is largest on GPQA-D and MedBullets,
where steps carry technical content that the extractor must restate and the
verifier must match precisely. \textsc{RC}'s benefit
therefore depends on a strong extractor, which accounts
for half of its cost.

\subsection{Different merging strategies}
\label{app:merge:strategies}

\begin{table}[htbp]
\centering
\small
\setlength{\tabcolsep}{2pt}
\begin{tabular}{@{}lccc@{}}
\toprule
 & \textbf{Cluster+verify} & \textbf{Pairwise} & \textbf{Embed.} \\
\midrule
SARA       & 86.8 & 84.8 & 85.8 \\
GPQA-D     & 60.6 & 59.7 & 55.1 \\
MedBullets & 70.5 & 71.8 & 65.4 \\
MuSR-MM    & 66.8 & 65.7 & 63.6 \\
MuSR-OP    & 52.7 & 52.7 & 50.8 \\
MuSR-TA    & 67.2 & 66.2 & 68.2 \\
FOLIO      & 84.7 & 83.7 & 83.7 \\
\midrule
All        & 69.9 & 69.1 & 67.5 \\

\midrule
Verifier calls / q & 16.6 & 1{,}247 & 0 \\
Merge tokens / q   & 17.0k & 867k & 0 \\

\bottomrule
\end{tabular}
\caption{Accuracy (\%) of \textsc{RC} under three merging strategies (seed 1234,
 majority voting: 68.9\%), and the cost of each strategy per question.}
\label{tab:merge}
\end{table}
\paragraph{Experiment:} We compare our cluster-then-verify merge with two alternatives, keeping
extraction and scoring fixed: (1)\emph{pairwise} verification, which sends every
cross-trace pair of same-type steps with cosine similarity
$\geq \tau$ to the verifier as a yes/no question and merges the confirmed pairs, and (1) \emph{embedding-only}
merging, which accepts every embedding cluster without LLM verification.

\newpage

\paragraph{Results: }Cluster-then-verify matches pairwise verification in accuracy (69.9\% vs.\
69.1\%, $p = 0.44$) with 75 times fewer verifier calls (16.6 vs.\ 1{,}247
per question) and 51 times fewer tokens.  Without LLM verification (only embedding),
accuracy falls to 67.5\%  (against
cluster-then-verify, $p = 0.007$). Only embedding merges steps that are
worded alike but state different claims (e.g. same computation with
different numbers). The verifier is therefore what makes the merged graph
useful, and clustering is what makes it affordable.

\subsection{Number of Traces}
\label{app:traces}

\begin{table}[htbp]
\centering
\small
\setlength{\tabcolsep}{3pt}
\begin{tabular}{@{}lcccccc@{}}
\toprule
 & \multicolumn{2}{c}{\textbf{4 traces}}
 & \multicolumn{2}{c}{\textbf{8 traces}}
 & \multicolumn{2}{c}{\textbf{12 traces}} \\
\cmidrule(lr){2-3}\cmidrule(lr){4-5}\cmidrule(lr){6-7}
\textbf{Dataset} & RC & MV & RC & MV & RC & MV \\
\midrule
SARA       & 84.9 & 83.5 & 85.7 & 84.9 & 86.8 & 85.7 \\
GPQA-D     & 53.5 & 53.0 & 58.1 & 56.6 & 60.6 & 58.1 \\
MedBullets & 67.4 & 67.9 & 67.9 & 67.4 & 70.5 & 66.3 \\
MuSR-MM    & 66.8 & 65.2 & 65.6 & 65.6 & 66.8 & 66.0 \\
MuSR-OP    & 52.0 & 52.3 & 54.3 & 52.3 & 52.7 & 53.1 \\
MuSR-TA    & 64.4 & 66.8 & 67.2 & 65.6 & 67.2 & 66.8 \\
FOLIO      & 82.3 & 81.8 & 82.8 & 85.2 & 84.7 & 85.2 \\
\midrule
All        & 67.5 & 67.4 & 68.9 & 68.3 & 69.9 & 68.9 \\
\midrule
Tied votes & \multicolumn{2}{c}{13.5\%} & \multicolumn{2}{c}{8.9\%}
 & \multicolumn{2}{c}{5.4\%} \\
\bottomrule
\end{tabular}
\caption{Accuracy (\%) of \textsc{RC} and majority voting with 4, 8 and 12
traces per question (one, two or three per generator; seed 1234), and the share
of questions on which the vote is tied.}
\label{tab:traces}
\end{table}

\paragraph{Experiment:} We vary the number of traces per question  keeping every pool balanced across models.  (seed 1234, all
1{,}619 questions).

\paragraph{Results: }Both methods improve with more traces (Table~\ref{tab:traces}). With four
traces, one per model, \textsc{RC} equals voting (67.5\% vs.\ 67.4\%); its
advantage grows to 0.6 points with eight and 1.0 with twelve. More traces give
the graph more corroboration to measure. A step stated by one of four traces
and one stated by three of four traces cannot be told apart as finely as with one of twelve vs eleven of twelve. The gain
is not uniform across benchmarks. Tt is consistent on SARA and GPQA-D, appears
only at twelve traces on MedBullets, and is absent or reversed on FOLIO and
MuSR-OP. Ties become rarer with more traces (13.5\% of questions with 4,
5.4\% with 12), so advantage at twelve traces does not come from
having more ties to break.

\newpage

\section{Diagnoses}
\label{app:diagnosis}

This appendix uses all three seeds and all questions.

\subsection{Why \textsc{RC} rarely overturns a majority}
\label{app:diagnosis:share}

 A step's attestation $\alpha(v)$
counts traces that state it, so a minority answer's
steps can have low $\alpha(v)$ as few traces argue for it. The score
thus amplifies the vote and keeps the order of clear majorities. A
minority answer can win only when vote is close and its steps are
corroborated more consistently across traces. This is where \textsc{RC}
changes the answer.

\subsection{Corroboration in answers \textsc{RC} changes}
\label{app:diagnosis:corrob}

We measure how much two traces agree as the Jaccard overlap of the merged
steps they state, and define the \emph{corroboration gap} of a question as the mean
overlap among traces of the correct answer minus that among traces of the
leading wrong answer. Where \textsc{RC} fixes a voting error, the correct traces agree more with each
other than the traces of the leading wrong answer (\emph{corroboration gap} +8.4 points). Where it
breaks a voting answer, the reverse holds (\emph{corroboration gap} $-6.4$). Over
questions with both RC and MV correct/ both wrong, the \emph{corroboration gap} is small (+0.8
$[+0.2, +1.4]$).


\subsection{FOLIO and the \emph{uncertain} label}
\label{app:diagnosis:folio}

\begin{table}[htbp]
\centering
\small
\setlength{\tabcolsep}{4pt}
\begin{tabular}{@{}lrcccc@{}}
\toprule
\textbf{Gold} & \textbf{$n$} & \textbf{RC} & \textbf{MV} & \textbf{Fixes}
 & \textbf{Breaks} \\
\midrule
\emph{true}      & 216 & 89.4 & 88.9 & 2 & 1 \\
\emph{false}     & 186 & 88.2 & 87.6 & 4 & 3 \\
\emph{uncertain} & 207 & 75.8 & 79.7 & 1 & 9 \\
\bottomrule
\end{tabular}
\caption{FOLIO accuracy (\%) by gold label over (question, seed) units, with
the units \textsc{RC} fixes and breaks relative to voting.}
\label{tab:folio}
\end{table}

\textsc{RC}'s loss on FOLIO comes entirely from questions whose gold label is
\emph{uncertain} (Table~\ref{tab:folio}): 9 of its 13 breaks against 1 of its 7
fixes. Lines concluding \emph{uncertain}
contain a similar number of Reasoning steps to lines concluding other options, and their traces share as many steps.
 When
\emph{uncertain} is the correct answer in a close vote, \textsc{RC} 
tends to prefer a definite answer. One possible reason for this is that arguments for \emph{uncertain} reason for uncertainty in different ways, which merge less often than shared
derivation steps for concrete answers.
\newpage

\section{Reproducibility}
\label{app:repro}

\paragraph{Models and sampling.} Generators: Gemma-4-E4B-it, Qwen3-8B
(thinking mode disabled), OLMo-3-7B-Instruct and Phi-4, sampled at temperature
0.6 and top-$p$ 0.95. Trace
$k$ of seed $s$ uses sampling seed $s + k$. The seeds are 1234, 2234 and 3234.
Extraction: Qwen3.8-27B (thinking disabled), temperature 0.3, top-$p$ 0.95,
seed 0. Merge verification: Qwen3.8-27B. Embeddings: all-mpnet-base-v2. The baselines use the
generators' sampling settings, with vLLM seed $s + 100\ell + i$ for layer or
round $\ell$ and model $i$.

\paragraph{Software and hardware.} All models run with vLLM with NVIDIA A100 40\,GB GPUs. Qwen3.8-27B and Phi-4 run  over two GPUs,
the other models on one. [Total compute: $\approx$ 260 GPU-hours.]

\paragraph{Data.} We use the full benchmarks: SARA (272 questions),
GPQA Diamond (198), the three MuSR domains (250, 256, 250) and FOLIO (203). For SARA, statute, case and hypothesis are taken as
separate fields from the LegalBench release. MedBullets has 304
questions, we use the 190 
that do not refer to a figure, as our models read text only.

\section{LLM Judge}
\label{app:judge}

\subsection{Protocol}
\label{app:judge:protocol}

\paragraph{Judge.} The judge is Claude Opus 5.5 (different model
family than generators and the extractor).

\paragraph{Items.} The judge is shown a the consensus subgraph 
 as a numbered list of typed steps in topological order. For the pairwise
comparison, the alternative is a \emph{random valid subgraph for the same answer}: an
unfolding HDAG from the consensus answer that picks a uniformly random
premise set at every \textsc{Or}, constrained to differ from the consensus in
at least one step. Every question had such a subgraph. Each pair is judged in both
order.

\paragraph{Analysis.} Let $y = 1$ if the judge picks the consensus. We fit
\[
\mathrm{logit}\,P(y = 1) = \beta_0 + \beta_\ell \log_2\frac{\ell_c}{\ell_a}
 + \beta_o\, o + u_q,  
\]
\[u_q \sim \mathcal{N}(0, \sigma^2),\]
with $\ell_c, \ell_a$  character lengths of the two explanations,
$o = \pm 1$ for consensus shown first or second. We report probability
of preferring consensus at equal length, averaged over $u_q$ and over
order.

\paragraph{Results: }We judged all 1{,}619 questions in both orders (3{,}238 judgments, all valid).
Table~\ref{tab:judge} gives the results per benchmark. The judge is consistent
across orders on 85\% of pairs and picks the first explanation in 50.8\% of
judgments ($\beta_o = 0.11$ $[-0.02, 0.23]$). It strongly favors length: each
doubling of the length ratio raises the log-odds of preference by
$\beta_\ell = 1.19$ $[0.74, 1.67]$. Preference at equal
length (0.61) exceeds raw preference (0.57).

\paragraph{Step overlap: }The two lines share 29\% of their merged steps at the median
(Table~\ref{tab:overlap}). The preference is larger when the
lines share fewer steps (0.60 below 0.25 vs.\ 0.55--0.57 above), on most
benchmarks individually as well, so near-identical pairs do not affect it the preference probability.

\begin{table}[htbp]
\centering
\small
\setlength{\tabcolsep}{2.5pt}
\begin{tabular}{@{}lccccc@{}}
\toprule
 & \textbf{Median} & \multicolumn{4}{c}{\textbf{Overlap bin: pairs / raw}} \\
\cmidrule(l){3-6}
\textbf{Dataset} & \textbf{overlap} & $<.25$ & $.25$--$.5$ & $.5$--$.75$
 & $\geq .75$ \\
\midrule
SARA       & 0.33 & 84/.67  & 105/.55 & 72/.54 & 11/.50 \\
GPQA-D     & 0.26 & 90/.65  & 83/.56  & 19/.63 & 6/.67 \\
MedBullets & 0.31 & 58/.68  & 93/.72  & 34/.63 & 5/.80 \\
MuSR-MM    & 0.26 & 114/.56 & 103/.54 & 30/.47 & 3/.83 \\
MuSR-OP    & 0.27 & 102/.56 & 126/.50 & 23/.52 & 5/.50 \\
MuSR-TA    & 0.25 & 123/.52 & 103/.51 & 17/.56 & 7/.64 \\
FOLIO      & 0.38 & 41/.60  & 92/.58  & 55/.54 & 15/.43 \\
\midrule
All        & 0.29 & 612/.60 & 705/.56 & 250/.55 & 52/.57 \\
\bottomrule
\end{tabular}
\caption{Jaccard overlap of the merged steps of the consensus and the random
same-answer line, and raw preference for the consensus by overlap bin.}

\vspace{-12pt}
\label{tab:overlap}
\end{table}
\section{Human Study: Design and Materials}
\label{app:human-study}

\subsection{Part 1}

\paragraph{Setup.} Eighteen graduate students recruited on a volunteer basis,  each evaluated six questions from MuSR, SARA and
MedBullets, two in each of the three formats described below, with the order
of formats shuffled across participants. The 18 questions were balanced by
the correctness of the majority answer (9 correct, 9 wrong), and in each
format a participant saw one question of each kind. Across participants,
every question appeared in each format exactly twice, giving 108 total trials.

\paragraph{Formats.}
Each question was presented in one of
three formats, all using reasoning from 4 LLMs:
\begin{itemize}
\item \textbf{Debate transcript.} A ReConcile-style multi-round
  discussion: 4 agents $\times$ 3 rounds, each response showing an
  answer and a stated confidence (0--1). The AI system's answer is the
  final-round vote weighted by confidence. 
\item \textbf{12 attempts + vote counts.} The 12 individual
  chain-of-thought attempts (4 LLMs $\times$ 3 samples), each showing
  its answer, with a vote tally. The AI system's answer is
  majority vote. 
\item \textbf{RC consensus graph.} The majority
  answer's consensus path and the minority answer's max-weight path. Each step is labelled
\textit{Fact}/\textit{Reasoning}/\textit{Conclusion} and annotated
  ``said by $n$ of 12 attempts,'' A
  small node--edge diagram and a node number key accompany the graph. No raw
  CoT is shown.
\end{itemize}
\vspace{-3pt}
All three formats give the same overall answer and the same questions and
options. Participants are told that the reasoning comes from ``an
AI system'' and never which format is the authors' method.

\vspace{-3pt}

\paragraph{Questions.}
For each item, participants answered:
 \textit{which main reasoning point most attempts relied on} (4~options: the central point,
the opposing side's point, a minor point, or a point nobody makes);
\textit{where the disagreeing attempts diverged in logic} (4~options: the real
divergence, a shared point, a majority-only point; ``Nobody
disagreed'').
They then rated understanding (1--5) and effort (1--5).
\vspace{-3pt}

\paragraph{Reference material.}
SARA items showed the case, the legal statement, and a verbatim statute
excerpt labelled with its tax code section, with a guide to reading
statutory numbering and an optional calculator. MuSR items showed a
collapsible summary and the full story. MedBullets items showed the
clinical vignette and answer options.
\vspace{-3pt}

\paragraph{Blind answer keys.}
Options and their keys were written by raters who saw only the
question, the 12 chain-of-thought attempts and the debate
transcript---never the consensus graph, the ground-truth label or any
participant responses. All 18 keys answerkeys were manually verified. Option numbers were seeded and shuffled across participants.

\paragraph{Significance} Reasoning comprehension was comparable across formats (graph 82\%, vote
82\%, debate 79\%; no pairwise difference significant). Graph questions
took the least effort, 0.9 points below vote on a 1--5 scale (Wilcoxon
$p = .015$), and tended to be fastest (2.9 min < vote,
$p = .06$).

\subsection{Part 2}
\paragraph{Extraction check.} 5 extracted reasoning steps per person
(30 unique steps, 3~raters each), shown alongside the likely source
passage. Annotators judged \textit{whether the source text says what the step
claims} and answered (yes / partly / no).

\subsection{Part 3}
\paragraph{Merge check. } 15 step pairs per person (90 unique pairs---48
merged, 21 verifier-rejected, 21 <$\tau$---3 raters each), order
of two steps randomised. Judge \textit{same claim / different
claims / unsure}.

\subsection{Part 3}
\paragraph{Survey. }: \textit{Asks for format preference.}

\section{Full Results}
\label{app:full}
\subsection{Accuracy per seed}
\label{app:full:seeds}

Table~\ref{tab:per-seed} gives accuracy for each seed. \textsc{RC} is above
majority voting on two seeds and level on the third (69.9, 69.7, 68.7 vs.\
68.9, 69.5, 68.7). Mixture-of-Agents varies most across seeds (68.7--72.1).

\subsection{Significance}
\label{app:full:sig}

A unit is a (question, seed) pair. For benchmark $d$ and seed $s$, let
$b_{d,s}$ count the units on which \textsc{RC} is correct and the other
method is not, and $c_{d,s}$ the reverse. We fit a hierarchical Bayesian
model that treats benchmarks as draws from a population and seeds as
replicates within a benchmark:
$\theta_d \sim \mathrm{Beta}(\mu\kappa, (1-\mu)\kappa)$,
$\theta_{d,s} \sim \mathrm{Beta}(\theta_d\lambda, (1-\theta_d)\lambda)$,
$b_{d,s} \sim \mathrm{Binomial}(n_{d,s}, \theta_{d,s})$ with
$n_{d,s} = b_{d,s} + c_{d,s}$, where $\mu \sim U(0,1)$ and $\kappa$ and
$\lambda$ have exponential priors with mean 50. Here $\theta$ is the
probability that \textsc{RC} is the correct one when exactly one method is
correct, and $P(\mu > 0.5)$ is the probability that \textsc{RC} is the
better method on a typical benchmark. It is 0.84 against majority voting,
0.20 against Mixture-of-Agents and 0.40 against ReConcile.





\subsection{Accuracy by vote margin}
\label{app:full:margin}

\begin{table}[htbp]
\centering
\small
\setlength{\tabcolsep}{2pt}
\newcommand{\ci}[2]{{\scriptsize[$#1$, $#2$]}}
\begin{tabular}{@{}lrrccc@{}}
\toprule
\textbf{Margin} & \textbf{Units} & \textbf{Share}
 & \textbf{MV} & \textbf{RC} & \textbf{Gain [95\% CI]} \\
\midrule
Tie ($0$) & 257 & 5.3\%
 & 40.9 & 49.4 & $+8.6$ \ci{+0.4}{+16.9} \\
Close ($(0, .25)$) & 659 & 13.6\%
 & 49.6 & 49.9 & $+0.3$ \ci{-2.3}{+3.0} \\
Mid ($[.25, .5)$) & 822 & 16.9\%
 & 55.0 & 54.5 & $-0.5$ \ci{-1.4}{+0.4} \\
Strong ($[.5, 1)$) & 1{,}954 & 40.2\%
 & 74.3 & 74.2 & $-0.1$ \ci{-0.2}{+0.0} \\
Unanimous ($1$) & 1{,}165 & 24.0\%
 & 87.4 & 87.4 & $+0.0$ \\
\bottomrule
\end{tabular}
\caption{Accuracy (\%) by the margin of the majority vote (vote share of the
top answer minus that of the runner-up), over all the 4{,}857 (question, seed) units.}
\label{tab:margin}
\end{table}

\textsc{RC} changes the voted answer almost only when the vote is close
(Table~\ref{tab:margin}). On tied votes, where voting can only pick one of the
tied answers at random, \textsc{RC} is correct 8.6 points more often. On unanimous votes the methods are
identical, and on strong majorities \textsc{RC} almost never
overrides vote.
\begin{table*}[t]
\centering
\small
\setlength{\tabcolsep}{4pt}
\begin{tabular}{@{}llcccccccc@{}}
\toprule
\textbf{Method} & \textbf{Seed}
 & \textbf{SARA} & \textbf{GPQA-D} & \textbf{MedB.}
 & \textbf{MuSR-MM} & \textbf{MuSR-OP} & \textbf{MuSR-TA}
 & \textbf{FOLIO} & \textbf{Avg} \\
\midrule
\textsc{RC} & 1234 & 86.8 & 60.6 & 70.5
 & 66.8 & 52.7 & 67.2 & 84.7 & 69.9 \\
 & 2234 & 87.1 & 59.1 & 68.4
 & 68.0 & 53.5 & 68.0 & 82.3 & 69.7 \\
 & 3234 & 86.4 & 55.6 & 70.0
 & 64.8 & 50.8 & 66.8 & 86.2 & 68.7 \\
\midrule
Majority vote & 1234 & 85.7 & 58.1 & 66.3
 & 66.0 & 53.1 & 66.8 & 85.2 & 68.9 \\
 & 2234 & 86.0 & 58.6 & 68.4
 & 66.8 & 52.7 & 68.0 & 85.2 & 69.5 \\
 & 3234 & 86.0 & 55.1 & 68.4
 & 67.2 & 50.0 & 68.0 & 85.7 & 68.7 \\
\midrule
Mixture-of-Agents & 1234 & 82.0 & 59.1 & 73.7
 & 68.4 & 48.8 & 64.4 & 86.7 & 68.7 \\
 & 2234 & 86.4 & 65.7 & 75.8
 & 72.4 & 48.0 & 71.2 & 86.7 & 72.1 \\
 & 3234 & 84.6 & 64.6 & 70.5
 & 75.6 & 44.5 & 67.2 & 85.2 & 70.2 \\
\midrule
ReConcile & 1234 & 82.4 & 59.1 & 72.6
 & 71.6 & 45.7 & 70.0 & 86.2 & 69.5 \\
 & 2234 & 85.7 & 60.6 & 72.6
 & 68.8 & 48.0 & 66.8 & 86.7 & 69.7 \\
 & 3234 & 86.4 & 61.1 & 72.6
 & 68.0 & 45.7 & 68.8 & 84.2 & 69.4 \\
\midrule
Gemma-4-E4B & 1234 & 85.9 & 58.1 & 64.6
 & 66.4 & 44.3 & 69.2 & 81.9 & 67.3 \\
 & 2234 & 85.8 & 58.6 & 64.9
 & 67.3 & 43.6 & 67.9 & 84.2 & 67.5 \\
 & 3234 & 84.4 & 57.6 & 66.1
 & 66.1 & 45.4 & 70.7 & 82.4 & 67.6 \\
\midrule
Qwen3-8B & 1234 & 77.9 & 47.1 & 62.5
 & 60.1 & 53.5 & 59.5 & 75.4 & 62.6 \\
 & 2234 & 80.5 & 44.3 & 63.0
 & 59.5 & 56.0 & 62.1 & 74.1 & 63.3 \\
 & 3234 & 81.1 & 44.8 & 63.7
 & 58.1 & 55.5 & 59.2 & 74.2 & 62.8 \\
\midrule
OLMo-3-7B & 1234 & 73.5 & 36.9 & 40.2
 & 58.3 & 46.9 & 53.7 & 62.9 & 54.2 \\
 & 2234 & 70.8 & 40.1 & 41.8
 & 57.6 & 47.1 & 55.5 & 63.7 & 54.6 \\
 & 3234 & 71.8 & 35.9 & 41.4
 & 59.6 & 45.7 & 54.5 & 65.2 & 54.3 \\
\midrule
Phi-4 & 1234 & 82.5 & 54.4 & 69.1
 & 68.5 & 47.9 & 63.7 & 78.8 & 66.5 \\
 & 2234 & 82.0 & 55.6 & 70.7
 & 67.5 & 48.0 & 65.9 & 78.3 & 66.9 \\
 & 3234 & 81.9 & 52.2 & 67.4
 & 68.1 & 44.1 & 65.7 & 80.8 & 65.8 \\
\bottomrule
\end{tabular}
\caption{Accuracy (\%) per seed. Single models: mean accuracy of their three
traces per question. }
\label{tab:per-seed}
\end{table*}

\section{Examples}
\label{app:examples}

\subsection{A consensus joined from several traces}
\label{app:examples:joined}

\begin{figure*}[t]
\centering
\includegraphics[width=\textwidth]{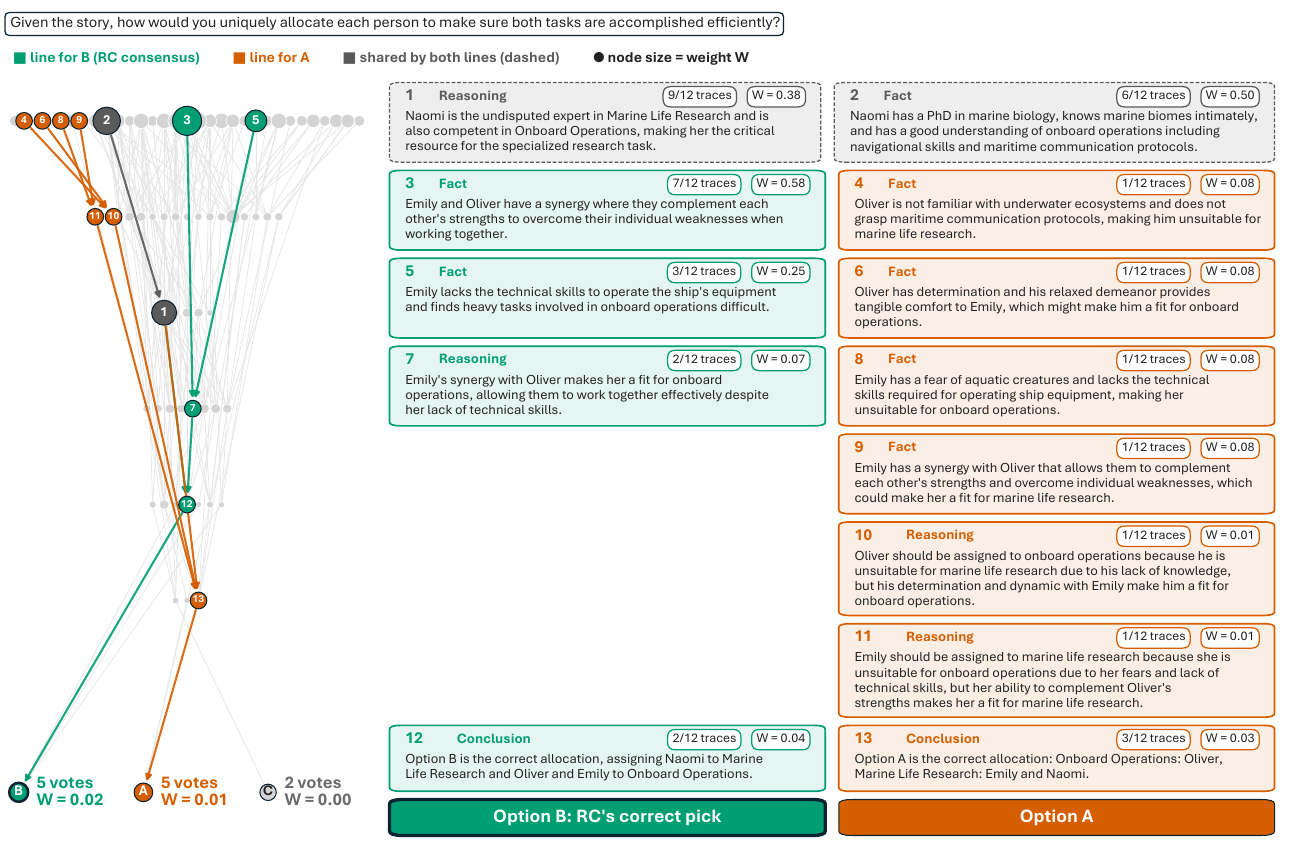}
\caption{MuSR-TA, seed 1234: assign Naomi, Emily and Oliver to two tasks. Votes
A 5, B 5, C 2; gold B. Both lines rest on Naomi's expertise (steps 1--2, 9 and 6
traces). The line for B (\textsc{RC}) justifies the roles of Emily and Oliver
with their complementary strengths (step 3, 7 traces), while the line for A
uses five steps stated by one trace each (4, 6, 8--11).}
\label{fig:ex-joined}
\end{figure*}

In Figure~\ref{fig:ex-joined}, no single trace states every step of the
consensus line; its steps come from traces of all four models, and step 3 is
stated by two traces that voted for A. Voting breaks the 5--5 tie at random;
\textsc{RC} answers B as it is corroborated more widely.

\subsection{A tie decided by corroboration}
\label{app:examples:tie}

\begin{figure*}[t]
\centering
\includegraphics[width=\textwidth]{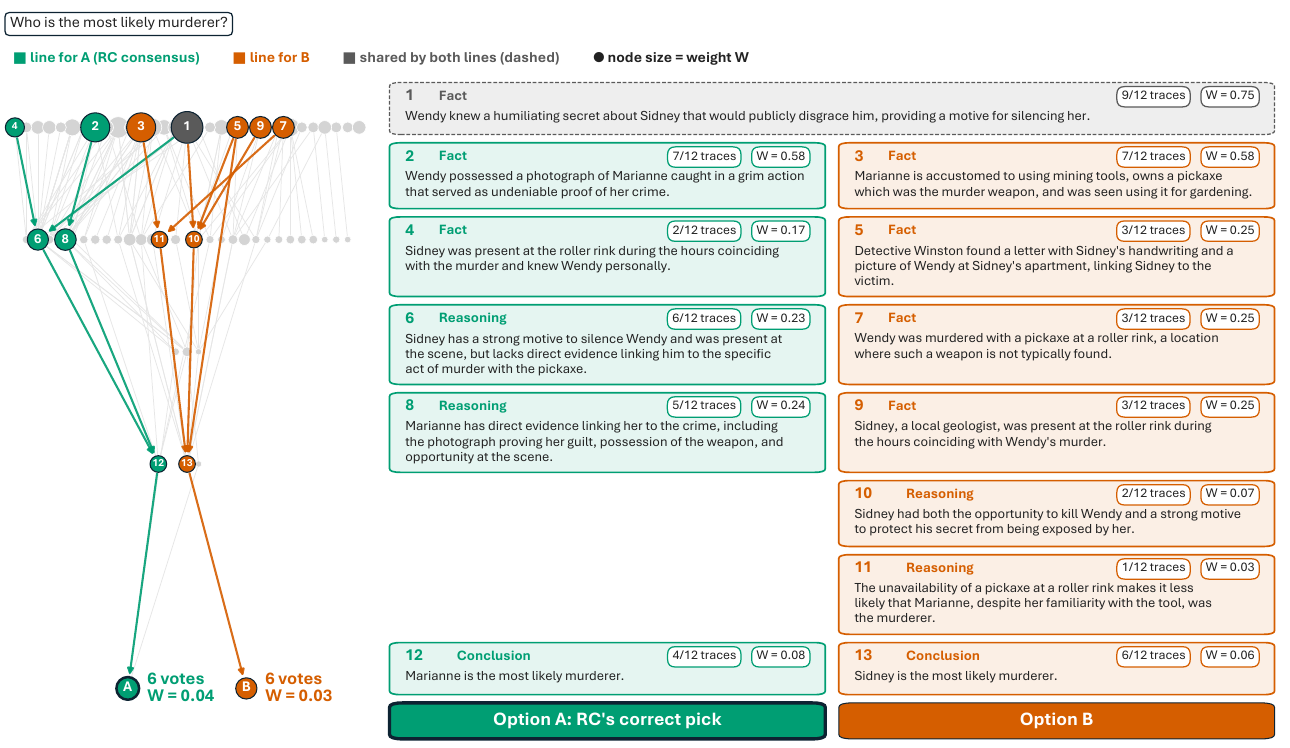}
\caption{MuSR-MM, seed 3234: who is the most likely murderer? Votes A
(Marianne) 6, B (Sidney) 6; gold A. The line for A rests on a photograph of
Marianne (step 2, 7 traces) and on the lack of direct evidence against Sidney
(step 6, 6 traces); the line for B rests on Sidney's motive (step 1, shared)
and reasoning steps stated by one or two traces (10, 11).}
\label{fig:ex-tie}
\end{figure*}

In figure \ref{fig:ex-tie}, both answers have evidence in the story, and the shared step (1) gives Sidney a
motive. The difference lies in the reasoning: the steps that weigh the evidence
for A are stated by five to seven traces, those for B by one or two.

\subsection{Reasoning the vote discards}
\label{app:examples:minority}

\begin{figure*}[t]
\centering
\includegraphics[width=\textwidth]{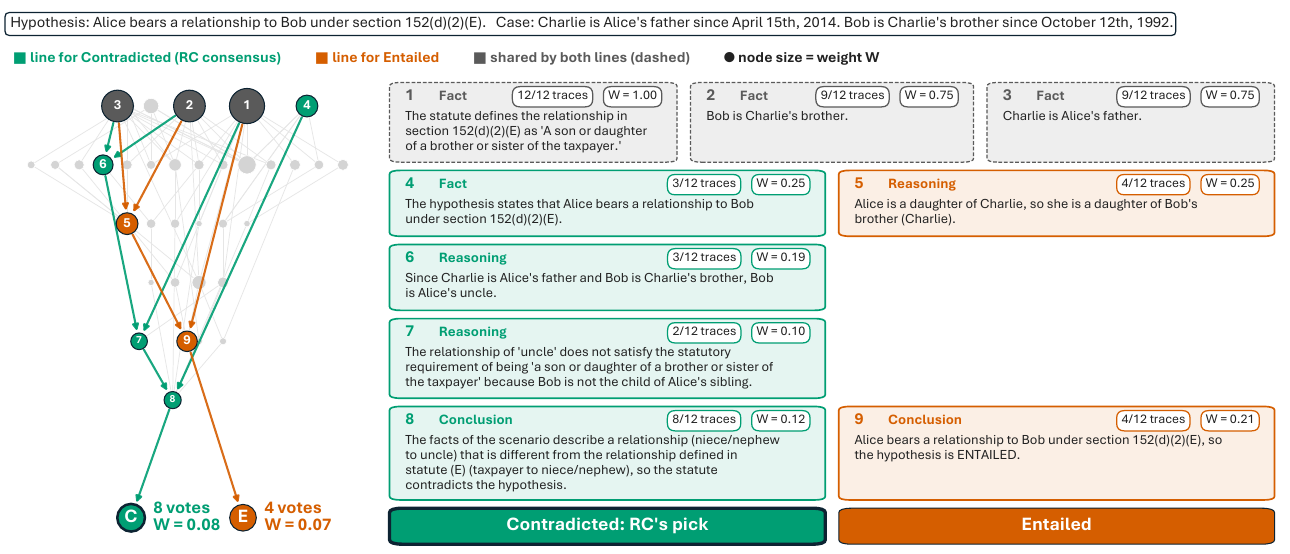}
\caption{SARA, seed 1234. Votes \emph{contradicted} 8, \emph{entailed} 4;
gold \emph{entailed}; both voting and \textsc{RC} answer \emph{contradicted}.
The lines share the statute and the two family facts (steps 1--3) and part at
one step: the majority infers that Bob is Alice's uncle (step 6, 3 traces),
placing Alice in the taxpayer's role; the minority infers that Alice is the
daughter of Bob's brother (step 5, 4 traces).}
\label{fig:ex-minority}
\end{figure*}

Figure~\ref{fig:ex-minority} shows a case where both methods are wrong but the
graph keeps the correct argument. A reader sees that the disagreement reduces
to the direction of the relationship, which the statute settles; the
minority's step is even stated by more traces than the majority's, though the
majority line wins on its conclusion and longer support.

\subsection{Different routes to the same answer}
\label{app:examples:routes}

\begin{figure*}[t]
\centering
\includegraphics[width=\textwidth]{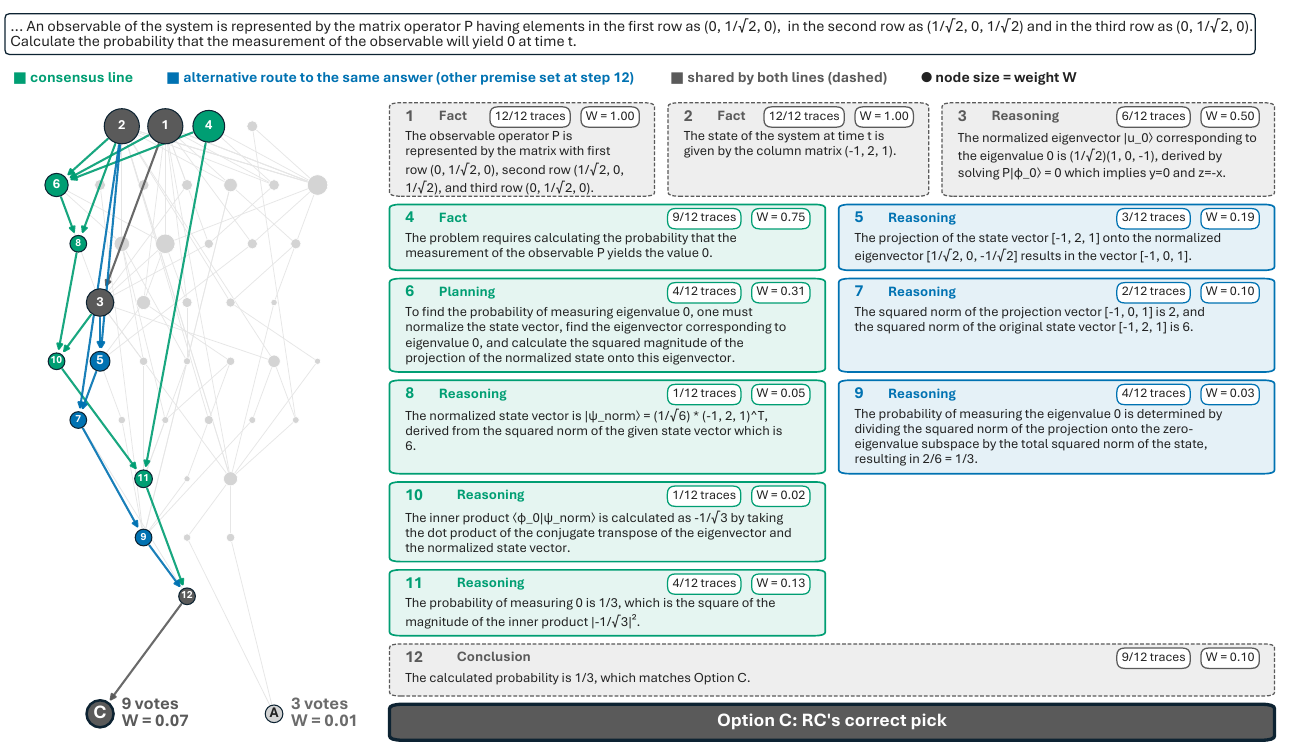}
\caption{GPQA-D, seed 2234: the probability that a measurement yields 0. Votes
C 9, A 3; gold C. Green: the consensus line computes the inner product of the
normalised state with the eigenvector of eigenvalue 0 and squares it (steps 8,
10, 11). Orange: an alternative route through the same graph projects the
unnormalised state onto the eigenvector and divides the squared norms, $2/6$
(steps 5, 7, 9). Both reach $1/3$, each stated by four traces at its final
step (11, 9).}
\label{fig:ex-routes-gpqa}
\end{figure*}

\begin{figure*}[t]
\centering
\includegraphics[width=\textwidth]{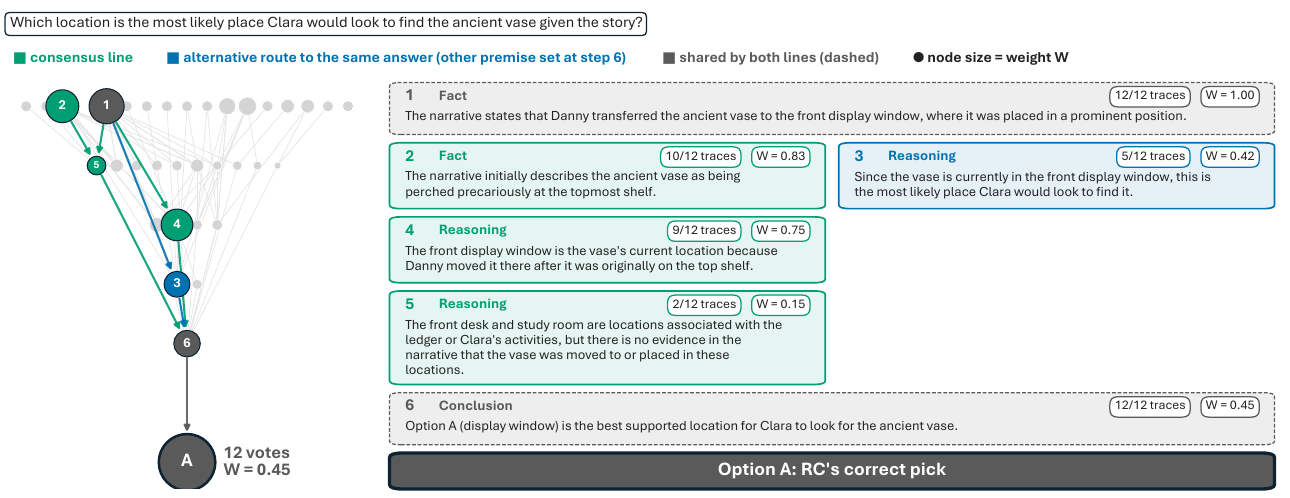}
\caption{MuSR-OP, seed 2234: where would Clara look for the vase? All 12 traces
answer A (the display window). Green: the consensus line argues from the vase's
current location (step 4, 9 traces) and rules out the other locations (step 5,
2 traces). Orange: alternative route argues directly that Clara would look
where vase is (step 3, 5 traces).}
\label{fig:ex-routes-op}
\end{figure*}

A step with several premise sets records different justifications of the
same claim. Choosing another premise set at such a step gives an alternative line to
the same answer. Figures~\ref{fig:ex-routes-gpqa} and~\ref{fig:ex-routes-op} show two
cases where the routes differ. In Figure~\ref{fig:ex-routes-gpqa}, the graph keeps two independent calculations
of the answer, each produced by several traces. A reader can check the answer
both ways, and their agreement is itself evidence. In
Figure~\ref{fig:ex-routes-op}, all traces agree on the answer, but they justify
it in two ways, supporting the answer or eliminating alternatives.

\subsection{Failure cases}
\label{app:examples:failures}

\begin{figure*}[t]
\centering
\includegraphics[width=\textwidth]{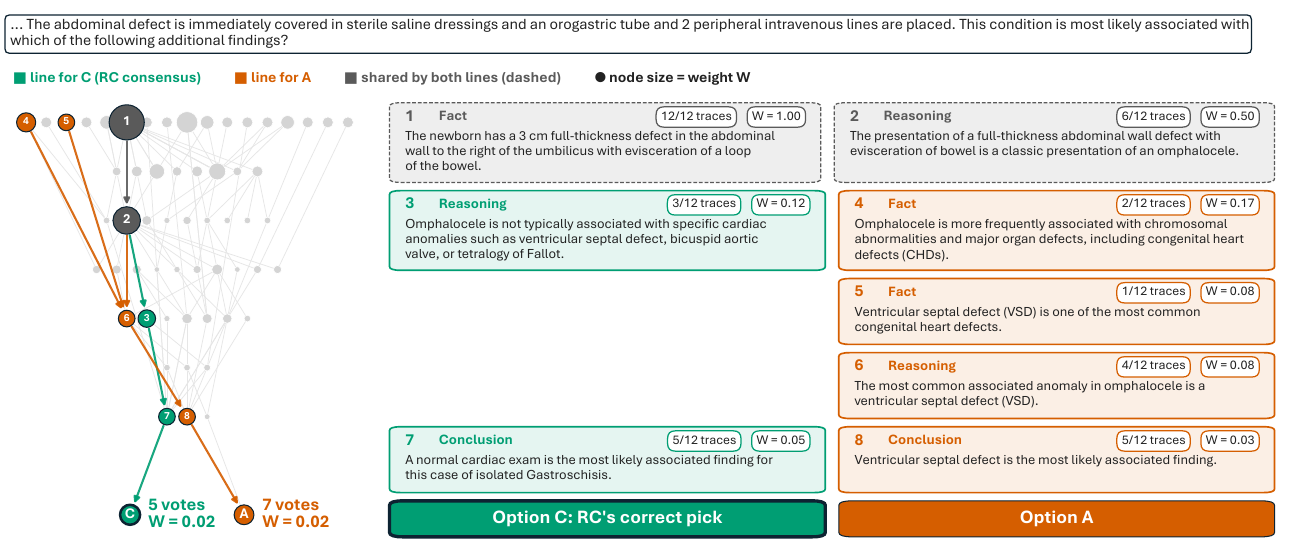}
\caption{MedBullets, seed 1234: a newborn with an abdominal wall defect to the
right of the umbilicus; which cardiac finding is most likely? Votes A
(ventricular septal defect) 7, C (normal exam) 5; gold C. \textsc{RC} answers
correctly, but its line calls the defect an omphalocele (step 2, 6 traces) and
concludes for a gastroschisis (step 7).}
\label{fig:ex-wrongstep}
\end{figure*}

\begin{figure*}[t]
\centering
\includegraphics[width=\textwidth]{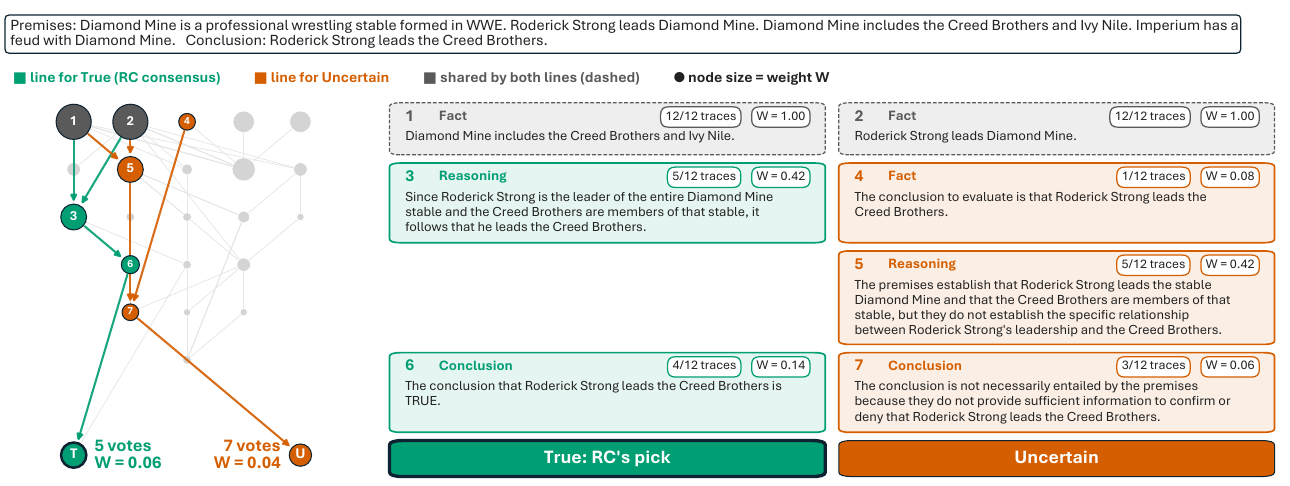}
\caption{FOLIO, seed 2234. Votes \emph{uncertain} 7, \emph{true} 5; gold
\emph{uncertain}; \textsc{RC} answers \emph{true}. Both lines rest on the same
premises (steps 1--2) and a central step stated by five traces (3, 5).}
\label{fig:ex-uncertain}
\end{figure*}

\noindent\textbf{A well-supported wrong step.} In
Figure~\ref{fig:ex-wrongstep}, \textsc{RC} is correct, but a step stated by
six traces calls the defect an omphalocele while the conclusion names
gastroschisis. Agreement made a wrong step look reliable; the graph shows
it, but the reader must catch it.

\noindent\textbf{An \emph{uncertain} majority overturned.} In
Figure~\ref{fig:ex-uncertain}, both lines are equally strong at their core,
but the \emph{uncertain} traces word their conclusion so differently that
only three of seven merge, against four of five for \emph{true}
(Appendix~\ref{app:diagnosis:folio})..